\documentclass{article}

\usepackage{microtype}
\usepackage{graphicx}
\usepackage{subfigure}
\usepackage{booktabs} 

\usepackage{hyperref}

\usepackage[numbers]{natbib}

\usepackage[accepted]{icml2023}

\usepackage{amsmath}
\usepackage{amssymb}
\usepackage{mathtools}
\usepackage{amsthm}

\usepackage[capitalize,noabbrev]{cleveref}

\theoremstyle{plain}

\theoremstyle{definition}

\theoremstyle{remark}

\usepackage[textsize=tiny]{todonotes}

\usepackage{times}
\usepackage{tabularx}
\usepackage{url}
\usepackage{multirow}
\Crefname{algocf}{Algorithm}{Algorithms}

\renewcommand{\algorithmicrequire}{\textbf{Input:}}
\renewcommand{\algorithmicensure}{\textbf{Output:}}

\usepackage{amsfonts}
\usepackage{wrapfig}
\usepackage{xcolor}
\usepackage{colortbl}

\newcommand{\Cc}{\mathcal{C}}

\newcommand{\Fc}{\mathcal{F}}

\newcommand{\Ic}{\mathcal{I}}

\newcommand{\Mc}{\mathcal{M}}

\newcommand{\Tc}{\mathcal{T}}

\newcommand{\Icb}{{\boldsymbol \Ic}}
\newcommand{\Fcb}{{\boldsymbol \Fc}}
\newcommand{\Ccb}{{\boldsymbol \Cc}}

\newcommand{\phib}{\boldsymbol \phi}

\newcommand{\0}{{\boldsymbol 0}}

\newcommand{\Fb}{{\boldsymbol F}}

\newcommand{\Ib}{{\boldsymbol I}}

\newcommand{\Pb}{{\boldsymbol P}}

\newcommand{\bb}{{\boldsymbol b}}

\newcommand{\fb}{{\boldsymbol f}}
\newcommand{\gb}{{\boldsymbol g}}

\newcommand{\nb}{{\boldsymbol n}}

\newcommand{\qb}{{\boldsymbol q}}

\newcommand{\ub}{{\boldsymbol u}}

\newcommand{\xb}{{\boldsymbol x}}

\newcommand{\pd}[2]{ \dfrac{\partial #1}{\partial #2}}
\newcommand{\pdflat}[2]{\tfrac{\partial #1}{\partial #2}}

\newcommand{\grad}{{\nabla}}
\newcommand{\gradb}{{\boldsymbol \nabla}}

\DeclareMathOperator{\trace}{\sf tr}

\newcommand{\figspace}{\vspace{-12pt}}

\newcommand{\RR}[1]{\mathbb{R}^{#1}}

\newcommand{\pdelhs}{\Fcb}
\newcommand{\TimeIntegration}{\Icb}
\newcommand{\TimeIntegrationAdv}{\Icb_{adv}}
\newcommand{\TimeIntegrationPro}{\Icb_{pro}}
\newcommand{\TimeIntegrationCor}{\Icb_{cor}}
\newcommand{\BoundaryConstraint}{\Ccb}
\newcommand{\pos}{\xb}

\newcommand{\continuousField}{{\fb}}
\newcommand{\continuousFieldArgs}[2]{\continuousField(#1,#2)}
\newcommand{\continuousFieldDiscrete}{{\fb}^{n}}
\newcommand{\continuousFieldDiscreteArg}[1]{\continuousFieldDiscrete(#1)}
\newcommand{\continuousFieldDiscretePlus}{{\fb}^{n+1}}

\newcommand{\continuousFieldDiscreteAlternative}{{\fb}^{k}}
\newcommand{\SetAllDiscreteTime}[1]{\{#1\}_{k=0}^{n+1}}
\newcommand{\SetAllDiscreteTimePrev}[1]{\{#1\}_{k=0}^{n}}

\newcommand{\nnweights}{\theta^{n}}
\newcommand{\nnweightsPlus}{\theta^{n+1}}
\newcommand{\continuousFieldNN}{{\fb_{\nnweights}}}
\newcommand{\discreteTime}{t_{n}}
\newcommand{\discreteTimePlus}{t_{n+1}}

\newcommand{\nnweightsAlternative}{\theta^{k}}
\newcommand{\continuousFieldNNAlternative}{{\fb_{\nnweightsAlternative}}}
\newcommand{\continuousFieldNNPlus}{{\fb_{\theta^{n+1}}}}

\newcommand{\nTemporal}{{T}}
\newcommand{\nSpatial}{{P}}
\newcommand{\temporalDiscretization}{\{\discreteTime\}_{n=0}^{\nTemporal}}

\newcommand{\spatialDomain}{\Omega}
\newcommand{\temporalDomain}{\Tc}

\newcommand{\dimensionIn}{m}
\newcommand{\dimensionOut}{d}

\DeclareMathOperator*{\argmin}{argmin}

\newcommand{\dt}{\Delta t}

\newcommand{\mlpwidth}{\beta}
\newcommand{\mlpdepth}{\alpha}

\newcommand{\advectQuantity}{u}
\newcommand{\advectSpeed}{a}
\newcommand{\advectQuantityDiscrete}{{\advectQuantity}^{n}}
\newcommand{\advectQuantityDiscretePlus}{{\advectQuantity}^{n+1}}

\newcommand{\fluidVelo}{\ub}
\newcommand{\press}{p}
\newcommand{\externalF}{\gb}
\newcommand{\fluidDen}{\rho_{f}}
\newcommand{\fluidVeloDiscrete}{{\fluidVelo}^{n}}
\newcommand{\fluidVeloPlus}{{\fluidVelo}^{n+1}}
\newcommand{\fluidVeloAdv}{\ub_{adv}}
\newcommand{\fluidVeloAdvPlus}{{\fluidVeloAdv}^{n+1}}
\newcommand{\pressPlus}{{\press}^{n+1}}

\newcommand{\sampleSet}{\Mc}

\newcommand{\defeq}{:=}

\newcommand{\discretePosSample}{{\xb^{j}}}
\newcommand{\sampleSetCardinality}{|\sampleSet|}
\newcommand{\sampleSetSummation}{\sum_{\pos\in\sampleSet\subset\spatialDomain}}

\newcommand{\posBT}{\pos_{backtrack}}
\newcommand{\squaredNorm}[1]{\|#1\|^2_2}

\newcommand{\deformMap}{\phib}
\newcommand{\deformGrad}{\Fb}
\newcommand{\solidDen}{\rho_{0}}
\newcommand{\pkstress}{\Pb}
\newcommand{\bodyforce}{\bb}
\newcommand{\externalforce}{\fb}
\newcommand{\energyDensity}{\Psi}

\newcommand{\firstlame}{\lambda}
\newcommand{\secondlame}{\mu}
\newcommand{\identity}{\Ib}
\newcommand{\singValues}{\Sigma}
\newcommand{\determinant}{\operatorname{det}}

\newcommand{\defomMapDiscrete}{{\deformMap}^{n}}
\newcommand{\defomMapDiscretePlus}{{\deformMap}^{n+1}}

\newcommand{\defomMapDotDiscrete}{{\dot{\deformMap}}^{n}}
\newcommand{\defomMapDotDiscretePlus}{{\dot{\deformMap}}^{n+1}}

\newcommand{\disretizationShape}{{N}}
\newcommand{\disretizationShapeDiscrete}{{\disretizationShape^i}}

\newcommand{\continuousFieldDiscreteSpace}{{\fb}^{n}_{i}}

\newcommand{\lrInit}{\textbf{lr}_0}
\newcommand{\lrMin}{\textbf{lr}_{\text{min}}}
\newcommand{\iterPatience}{\textbf{iter}_{\text{p}}}
\newcommand{\iterMax}{\textbf{iter}_{\text{max}}}
\newcommand{\timeAvg}{\textbf{t}_{\text{avg}}}

\begin{document}

\setlength{\abovedisplayskip}{5pt}
\setlength{\belowdisplayskip}{5pt}

\twocolumn[
\icmltitle{Implicit Neural Spatial Representations for Time-dependent PDEs}
\icmltitlerunning{Implicit Neural Spatial Representations for Time-dependent PDEs}



\icmlsetsymbol{equal}{*}

\begin{icmlauthorlist}
\icmlauthor{Honglin Chen}{equal,Columbia}
\icmlauthor{Rundi Wu}{equal,Columbia}
\icmlauthor{Eitan Grinspun}{UofT}
\icmlauthor{Changxi Zheng}{Columbia}
\icmlauthor{Peter Yichen Chen}{CSAIL,Columbia}

\end{icmlauthorlist}

\icmlaffiliation{Columbia}{Department of Computer Science, Columbia University, New York, USA}
\icmlaffiliation{CSAIL}{Computer Science and Artificial Intelligence Laboratory, Massachusetts Institute of Technology, Cambridge, USA}
\icmlaffiliation{UofT}{Department of Computer Science, University of Toronto, Toronto, Canada}

\icmlcorrespondingauthor{Honglin Chen}{honglin.chen@columbia.edu}
\icmlcorrespondingauthor{Rundi Wu}{rundi.wu@columbia.edu}

\icmlkeywords{Machine Learning, ICML}

\vskip 0.3in
]



\printAffiliationsAndNotice{\icmlEqualContribution} 

\begin{abstract}
Implicit Neural Spatial Representation (INSR) has emerged as an effective representation of spatially-dependent vector fields. This work explores solving time-dependent PDEs with INSR. Classical PDE solvers introduce both temporal and spatial discretizations. Common spatial discretizations include meshes and meshless point clouds, where each degree-of-freedom corresponds to a location in space. While these \emph{explicit} spatial correspondences are intuitive to model and understand, these representations are not necessarily optimal for accuracy, memory usage, or adaptivity. Keeping the classical temporal discretization unchanged (e.g., explicit/implicit Euler), we explore INSR as an alternative spatial discretization, where spatial information is \emph{implicitly} stored in the neural network weights. The network weights then evolve over time via time integration. Our approach does \emph{not} require any training data generated by existing solvers because our approach is the solver itself. We validate our approach on various PDEs with examples involving large elastic deformations, turbulent fluids, and multi-scale phenomena. While slower to compute than traditional representations, our approach exhibits higher accuracy and lower memory consumption. Whereas classical solvers can dynamically adapt their spatial representation only by resorting to complex remeshing algorithms, our INSR approach is intrinsically adaptive. By tapping into the rich literature of classic time integrators, e.g., operator-splitting schemes, our method enables challenging simulations in contact mechanics and turbulent flows where previous neural-physics approaches struggle. Videos and codes are available on the project page.\footnote{Project webpage: \hyperlink{http://www.cs.columbia.edu/cg/INSR-PDE/}{http://www.cs.columbia.edu/cg/INSR-PDE/}}
\end{abstract}
\section{Introduction}
Implicit neural spatial representation (INSR) \cite{park2019deepsdf, xie2021neural} parameterizes a spatially-dependent vector field with a neural network. It has been proven to be instrumental in computer graphics and vision applications, including volumetric rendering \cite{mildenhall2020nerf}, 3D reconstruction \cite{mescheder2019occupancy}, signal processing \citep{du2021learning}, and geometry processing \cite{yang2021geometry}. While existing works have mostly focused on static representations, this work aims to explore these neural representations for dynamic simulations where the vector fields evolve over time. In particular, we explore a fundamental class of physics simulation tasks governed by partial differential equations (PDEs) with both \emph{spatial} and \emph{temporal} dependence,
\begin{align}
\begin{split}
    \pdelhs(\continuousField, \gradb\continuousField, \gradb^2\continuousField, \ldots, \dot{\continuousField}, \ddot{\continuousField}, \ldots) = \0, \\ \continuousFieldArgs{\pos}{t}: \spatialDomain \times \temporalDomain \to \RR{\dimensionOut} \ ,    
\end{split}
\label{thepde}
\end{align}
where $\spatialDomain\in\RR{\dimensionIn}$ and $\temporalDomain\in\RR{}$ are the spatial and temporal domains, respectively. Examples include the Navier-Stokes equations for fluid dynamics and the elastodynamic equation for solid mechanics.

To computationally simulate these problems, classical methods introduce both \emph{spatial} and \emph{temporal} discretizations. On the one hand, \emph{temporal} discretization breaks down the entire temporal range into a finite number of time steps $\temporalDiscretization$, where $\nTemporal$ is the number of temporal discretization samples, and $\dt=\discreteTimePlus-\discreteTime$ is the time step size. The solution to \Cref{thepde} then becomes a list of spatially-dependent vector fields: $\{\continuousFieldDiscreteArg{\pos}\}_{n=0}^{\nTemporal}$. Classical methods then sequentially integrate from one time step ($n$) to the next ($n+1$), using a wide range of time integrators, such as explicit/implicit Euler \citep{ascher1998computer}. On the other hand, \emph{spatial} discretization represents these spatially-dependent vector fields $\continuousFieldDiscreteArg{\pos}$ using grids, meshes, or point clouds (meshless particles). For example, the grid-based linear finite element method (FEM) \citep{hughes2012finite} defines a shape function $\disretizationShapeDiscrete$ on each grid node and represents the spatially dependent vector field as $\continuousFieldDiscreteArg{\pos}=\sum_{i=1}^{\nSpatial}\continuousFieldDiscreteSpace\disretizationShapeDiscrete$, where $\nSpatial$ is the number of spatial samples.

While widely adopted in scientific computing applications, these classic \emph{spatial} discretizations are not without drawbacks:
\begin{enumerate} 
    \item Spatial discretization errors abound in fluid simulations as artificial numerical diffusion \citep{lantz1971quantitative}, dissipation \citep{fedkiw2001visual}, and viscosity \citep{roache1998fundamentals}. These errors also appear in solid simulations as inaccurate collision resolution \citep{muller2015air} and numerical fractures \citep{sadeghirad2011convected}.
    \item Memory usage spikes with the number of spatial samples $\nSpatial$ \citep{museth2013vdb}.
    \item Adaptive meshing \citep{narain2012adaptive} and data structures \citep{setaluri2014spgrid} can reduce memory footprints but are often computationally expensive and challenging to implement.
\end{enumerate}

In this work, we ask: \emph{what are the (dis)advantages of replacing a classical numerical method's spatial discretization with INSR while \underline{keeping intact} the time integrator?} Unlike traditional representations that \emph{explicitly} discretize the spatial vector via spatial primitives (e.g., points), INSRs \emph{implicitly} encode the field through neural network weights. 
In other words, the field is parameterized by a neural network (typically multilayer perceptrons), i.e., $\continuousFieldDiscrete(\pos)=\continuousFieldNN(\pos)$ with $\nnweights$ being the network weights. As such, the memory usage for storing the spatial field is independent of the number of spatial samples, but rather it is determined by the number of neural network weights. We show that under the same memory constraint, INSRs indeed achieve higher accuracy than traditional discrete representations. Furthermore, INSRs are adaptive by construction \citep{xie2021neural}, allocating the network weights to resolve field details at \emph{any} spatial location without changing the network architecture.

We emphasize that our contribution is orthogonal to the choice of \emph{temporal} discretization. Notably, our INSR approach works with various classic time integrators, including explicit/implicit/midpoint Euler, variational time integrators \citep{kane2000variational}, and even operator splitting schemes \citep{chorin1968numerical}. Indeed, our focus contrasts with other neural approaches, e.g., physics-informed neural network (PINN) \citep{raissi2019physics}. Whereas prior work has focused on overcoming low data availability, efficiently solving inverse problems, or addressing high-dimensional problems, our unique focus is on exploring incorporating various existing classical time integrators. This ability to leverage classic time integrators is particularly effective in highly nonlinear problems, e.g., turbulence, where previous neural-PDE approaches struggle, e.g., PINN.

In summary, we make the following contributions:\vspace{-2mm}
\begin{itemize}
    \item We present INSR as an alternative spatial representation for various time-dependent physics simulation problems.\vspace{-2mm}
    \item Compared to the classic grid, mesh, point cloud (meshless) spatial representations, our INSR approach trades wall-clock runtime in favor of three benefits: lower discretization error, lower memory usage, and built-in adaptivity.\vspace{-2mm}
    \item Utilizing various classic time integrators, including variational time integrators and operator splitting schemes, INSR-based simulations capture challenging cases in contact mechanics and turbulent flows where previous neural-physics approaches fail.\vspace{-2mm}
\end{itemize}

\begin{figure}[t]
  \centering
  \includegraphics[width=0.99\linewidth]{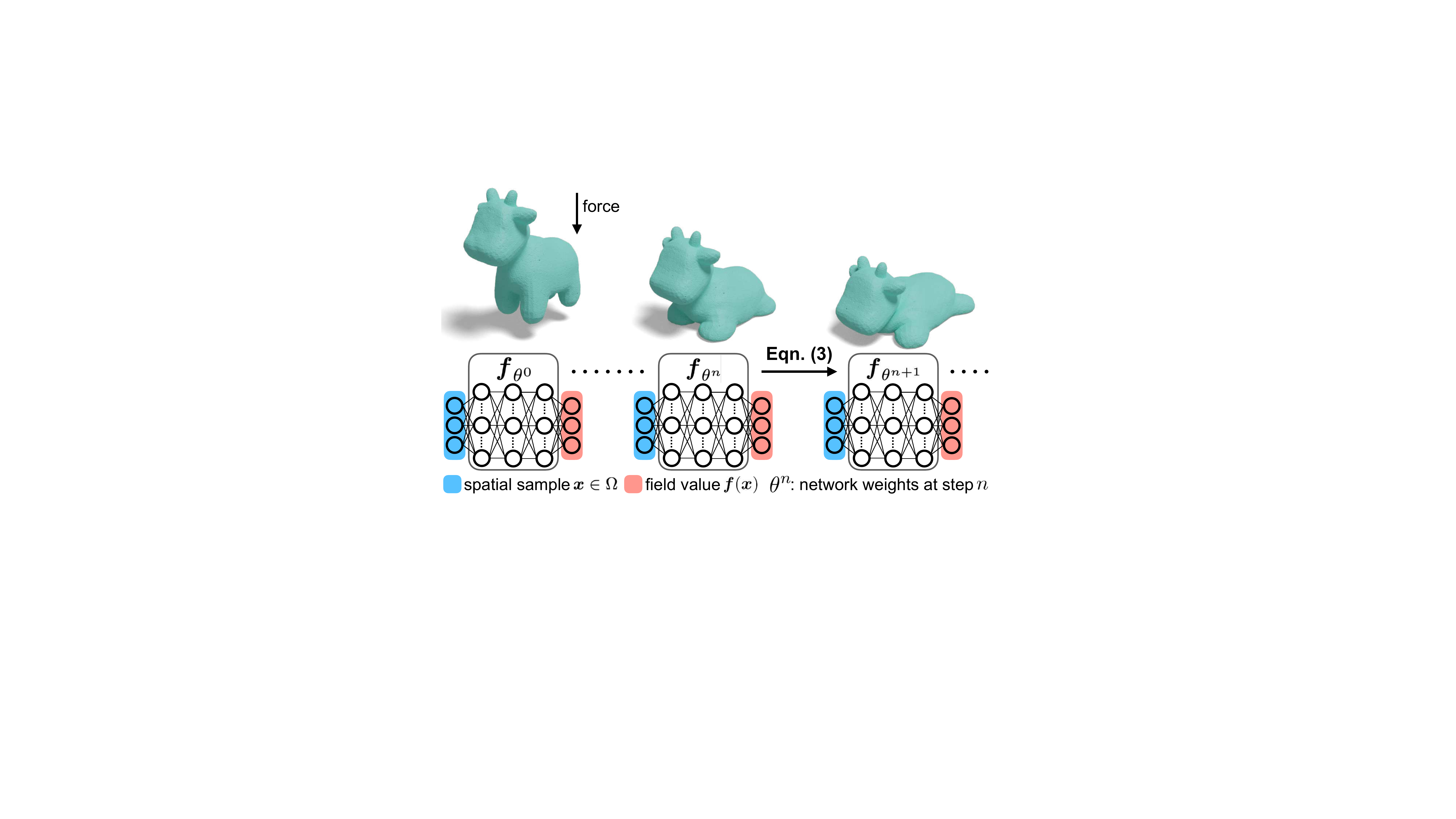}
  \vspace{-10pt}
    \caption{\textbf{Evolve neural field over time.}
    We represent the field of interest using a neural network $\continuousFieldNN$,
    whose weights $\nnweights$ are updated at each time step via optimization-based time integration (\Cref{eqn:opti-time-inte-nn}).
    In this case, the spatial domain $\spatialDomain$ is the volume encompassed by the undeformed object, and the represented field $\continuousField$ is the deformation map. The governing PDE is the elastodynamic equation.
    }
  \figspace
  \label{fig:opti-time-inte-nn}
\end{figure}
\vspace{-10pt}
\section{Related Works}
\textbf{Implicit Neural Spatial Representation} (INSR) uses neural networks to parameterize spatially-dependent functions \citep{chen2019learning,mescheder2019occupancy, mildenhall2020nerf,dupont2022data,chen2021model,pan2022neural,chen2022crom}, e.g., a signed-distance-field, where the input is an arbitrary spatial location and the output is its distance to the surface \cite{park2019deepsdf}. The nonlinear neural network's enormous expressivity makes INSR more accurate than its classic mesh-based and meshless counterparts under the same memory constraint. For example, with the same number of neural network weights as the number of mesh vertices (or meshless particles), INSR-SDF captures more geometric details than a triangle mesh \citep{takikawa2021neural}. 
Indeed, memory consumption of traditional representations scales poorly with
spatial resolutions. Adaptive discretizations can reduce memory, but their
generations are expensive. By contrast, neural representations are adaptive by
construction and can use their representation capacities at arbitrary locations
of interest without memory increases or data structure alterations \citep{xie2021neural}. 
Because of the above-mentioned advantages, researchers have used INSR for many other applications such as image processing \citep{chen2021learning,shaham2021spatially}, 3D reconstruction \citep{wang2021neus,yariv2020multiview}, generative modeling \citep{schwarz2020graf,wu2022learning,chan2022efficient} and geometry processing \citep{yang2021geometry,sharp2022spelunking}.
Besides, time-dependent implicit representations 
have also been explored for capturing scene dynamics \citep{park2021nerfies,park2021hypernerf}.

Regarding PDE applications, INSR has been successively deployed in strictly spatially dependent PDEs, including elastostatics \citep{zehnder2021ntopo}, elliptic PDEs \citep{chiaramonte2013solving}, and geometry processing \citep{yang2021geometry}. For time-dependent PDEs, \citet{mehta2022level} propose a framework for evolving INSR weights over time. However, their approach specializes in level sets. \citet{du2021evolutional,bruna2022neural} also evolve INSR's network weights over time, with the goal of removing PINN's limited time range as well as solving high-dimensional problems where meshing is impossible. By contrast, our work focuses on low-dimensional problems that heavily rely on classical FEM methods and explore the INSR solver as a more accurate, memory-efficient, and adaptive alternative.

\vspace{-10pt}
\paragraph{Machine Learning (ML) for PDEs} One line of ML-PDE works train on data from classical solvers (or real experiments), e.g., graph neural network (GNN) approaches \citep{sanchez2020learning,pfaff2020learning}, neural operator approaches \citep{li2020fourier,li2020neural}, and DeepONet \citep{lu2019deeponet}. After training, these methods solve a new problem faster than the solver on which it was trained. However, these methods typically do not generalize to arbitrary initial/boundary conditions, material parameters, or geometries that are drastically different from the ones presented in the training data \citep{wang2021long}. Another line of ML-PDE works does not require training data from classical solvers at all since these methods are \emph{solvers} themselves, i.e., given the PDE and initial/boundary conditions, these methods can directly solve the PDE just like the classical solvers. These methods usually employ a physics-informed loss term (e.g., $\|\pdelhs\|^2$) that incorporates the governing-PDE \citep{raissi2019physics,wandel2020learning,wang2021learning}. Our work also belongs to this ``solver-type''.

Previous physics-informed approaches generally treat the temporal domain as a continuous variable, and the PDE loss term would incorporate the entire spatiotemporal-dependent PDE ($\pdelhs$). \citet{krishnapriyan2021characterizing} break down the temporal domain into several subdomains and obtain better long-term temporal integration. However, the time variable is still treated as a continuous variable within each subdomain. By contrast, our approach discretizes the temporal domain just like the classical solvers. While this explicit temporal discretization can address the long-term prediction limitation of PINN, it is not the primary goal of this work. Instead, we show that by tapping into the rich literature of classical time integration schemes, we can model challenging problems in contact mechanics and turbulent flows where previous neural-PDE approaches struggle. \citet{raissi2019physics,wessels2020neural} also explore a time-discrete approach, but their methods specialize in Runge-Kutta schemes, while our general formulation supports a wide range of classical time integrators, including operator-splitting schemes.
\section{Method: Time Integration on Neural Spatial Representations}
Our goal is to solve time-dependent PDEs on neural-network-based spatial
representations. In \Cref{sec:nn-rep}, we first discuss representing spatial
vector fields with neural networks. Afterward, we will describe how to time-step the network weights with classic time integrators.

\subsection{Neural Networks as Spatial Representations}
\label{sec:nn-rep}
We parameterize each time-discretized spatial vector field with a neural network: $\continuousFieldDiscrete=\continuousFieldNN$, where $\nnweights$ are the neural network weights at time $\discreteTime$.
Specifically, the field quantity at an arbitrary spatial location $\pos \in \spatialDomain$ can be queried via network inference $\continuousFieldNN(\pos)$.

Traditional representations \emph{explicitly} discretize the spatial vector
field using primitives such as meshes. These primitives
\emph{explicitly} correspond to spatial locations due to their compactly
supported basis functions \citep{hughes2012finite}. By contrast, INSRs \emph{implicitly} encode the vector field via neural network
weights, and each weight affects the vector field globally. Such global
support is also an attribute of spectral methods
\citep{canuto2007spectral1,canuto2007spectral2}. Compared to spectral methods, our approach does not need to know the required complexity ahead of time in order to determine the ideal basis functions \citep{xie2021neural}. INSR automatically optimizes its parameters to where field detail is present.

Whereas memory consumption of \emph{explicit} representations scales poorly with the number of spatial samples, memory consumption for
INSR is independent of the number of spatial
samples. Instead, memory usage (for storing the vector field) is determined by the number
of network weights.

\vspace{-10pt}
\paragraph{Network Architecture}
Following the INSR literature, we adopt SIREN \citep{sitzmann2020implicit} (MLP with sinusoidal activation) as our network architecture for its accuracy and quick convergence speed advantages. Each MLP has a total of $\mlpdepth$ hidden layers, each layer of width $\mlpwidth$.
The specific choice of these hyper-parameters is described in \Cref{sec:experiments}.

\vspace{-10pt}
\paragraph{Spatial Gradients}
Classic spatial representations compute spatial gradients via basis functions. Higher-order gradients require higher-order basis functions. By contrast, INSR is $C^{\infty}$ by construction. We evaluate the gradients via computation-graph-based auto-differentiation with respect to the input (not the weights).

\subsection{Time integration}
\label{sec:time-stepping}
Given previous-time spatial vector fields $\{\continuousFieldDiscreteArg{\pos}\}_{k=0}^{n}$, we can compute the next time-step ($\discreteTimePlus$) by solving an optimization problem:
\begin{small}
\begin{align}
\begin{split}
    \continuousFieldDiscretePlus = \argmin_{\continuousFieldDiscretePlus} \sampleSetSummation\TimeIntegration(&\dt,\SetAllDiscreteTime{\continuousFieldDiscreteAlternative(\pos)},\SetAllDiscreteTime{\grad\continuousFieldDiscreteAlternative(\pos)},\\
    &\SetAllDiscreteTime{\grad^2\continuousFieldDiscreteAlternative(\pos)}, \ldots) \ .
    \label{eqn:opti-time-inte}
\end{split}
\end{align}
\end{small}

For example, the classic explicit/implicit Euler methods can be formulated in this form \citep{kharevych2006geometric} as well as variational time integrators derived from Hamilton's principle \citep{kane2000variational}. Operator-splitting style integrators \citep{chorin1968numerical} also weave seamlessly into this formulation by solving multiple optimization problems. Note that the particular choice of the objective function $\TimeIntegration$ depends on the PDE of interest and the time integrator choice.

This optimization formulation applies to \emph{any} spatial representation. It has been explored thoroughly for classic spatial discretizations \citep{batty2007fast,bouaziz2014projective,gast2015optimization}, which is defined over a finite number of the spatial integration samples $\sampleSet \defeq \{\discretePosSample \in \spatialDomain \ |\  1\leq j \leq \sampleSetCardinality\}$, e.g., grids. 

Applying this formulation to a neural spatial representation, we optimize for
\begin{small}
\begin{align}
\begin{split}
    \nnweightsPlus = \argmin_{\nnweightsPlus} \sampleSetSummation\TimeIntegration(&\dt,\SetAllDiscreteTime{\continuousFieldNNAlternative(\pos)},\SetAllDiscreteTime{\grad\continuousFieldNNAlternative(\pos)},\\
    &\SetAllDiscreteTime{\grad^2\continuousFieldNNAlternative(\pos)},\ldots) \ 
\end{split}
    \label{eqn:opti-time-inte-nn}
\end{align}
\end{small}

where $\SetAllDiscreteTimePrev{\nnweightsAlternative}$ are the (fixed, not variable) neural network weights from previous time steps.
\Cref{fig:opti-time-inte-nn} illustrates our time integration process, and \Cref{ti_algo} provides the corresponding pseudocode. 
In all the examples presented in this work, we solve this time-integration optimization problem via Adam \citep{kingma2014adam}, a first-order stochastic gradient descent method.

\vspace{-10pt}
\paragraph{Spatial Sampling}
\label{para:sampling}
\emph{Explicit} spatial representations (e.g., tetrahedra mesh) are often tied to a particular spatial sampling; remeshing is sometimes possible but can also have drawbacks, especially in higher dimensions \citep{alliez2002interactive,narain2012adaptive}. By contrast, \emph{implicit} spatial representations allow for arbitrary spatial sampling by construction  (\Cref{eqn:opti-time-inte-nn}). Following \citet{sitzmann2020implicit}, we dynamically sample $\sampleSet$ during optimization. For every gradient descent iteration in every time step, we use a stochastic sample set $\sampleSet$ from the spatial domain $\spatialDomain$; $\sampleSet$ corresponds to the ``mini-batch'' in stochastic gradient descent, with batch size $\sampleSetCardinality$.

\paragraph{Boundary Condition}
 PDEs are typically accompanied by spatial (e.g.,  Dirichlet or Neumann) boundary conditions, which we formulate as additional penalty terms in the objective
\Cref{eqn:opti-time-inte-nn},
\begin{small}
\begin{align}
\begin{split}
    \nnweightsPlus = \argmin_{\nnweightsPlus} &\sampleSetSummation\TimeIntegration(\dt,\SetAllDiscreteTime{\continuousFieldNNAlternative(\pos)},\SetAllDiscreteTime{\grad\continuousFieldNNAlternative(\pos)}, \\
    \ldots) 
	+\lambda &\sum_{\pos^b\in\sampleSet^b\subset\partial\spatialDomain} \BoundaryConstraint(\continuousFieldNNPlus(\pos^b),\grad\continuousFieldNNPlus(\pos^b),\ldots) \ ,
\end{split}
\label{eqn:opti-time-inte-nn-bc}
\end{align}
\end{small}
where $\lambda$ is the weighting factor and $\partial\spatialDomain$ is the boundary of the spatial domain.
The particular choice of the boundary constraint function $\BoundaryConstraint$ depends on the problem of interest.

\begin{algorithm}[t]
\caption{Time integration}
\label{ti_algo}
\begin{small}
\begin{algorithmic}[1]
\REQUIRE 
    $\text{initial network weights } \theta^0, \text{timestep size } \dt,$\\
    $\text{number of timesteps } N, \text{time integrator } \TimeIntegration, \text{spatial domain } \spatialDomain$
\STATE $n\gets 0$
\WHILE{$n < N$}
\STATE $\nnweightsPlus \gets \nnweights$
    \WHILE{not converged}{
    		\STATE randomly sample $\sampleSet\subset\spatialDomain$
            \STATE  
            {
            \scriptsize   $L_{\nnweightsPlus}=\sum\limits_{\pos\in\sampleSet}\TimeIntegration(\dt,\SetAllDiscreteTime{\continuousFieldNNAlternative(\pos)},\SetAllDiscreteTime{\grad\continuousFieldNNAlternative(\pos)},\ldots)$
            }
    		\STATE $\nnweightsPlus \gets \nnweightsPlus - \alpha\nabla L_{\nnweightsPlus}$
    	}
    \STATE $n \gets n + 1$
    \ENDWHILE
\ENDWHILE
\end{algorithmic}
\end{small}
\end{algorithm}

\paragraph{Initial Condition}
The neural network is initialized using the given initial condition, i.e., the field value at time $t=0$, by optimizing
\begin{align}
	\theta^0=\argmin_{\theta^0} \sampleSetSummation ||\continuousField_{\theta^0}(\pos) - \hat{\continuousField^0}(\pos) ||_2^2 \ ,
\end{align}
where $\hat{\continuousField^0}$ is the given  initial condition.
Similar to \Cref{eqn:opti-time-inte-nn}, we solve this optimization problem using Adam \citep{kingma2014adam} and stochastically sample $\sampleSet$ at each gradient descent iteration.
\section{Experiments}
\label{sec:experiments}
In this section, we evaluate our method on three classic time-dependent PDEs: the advection equation, the incompressible Euler equations, and the elastodynamic equation. 

\begin{figure}[t]
  \begin{center}
  \includegraphics[width=\linewidth]{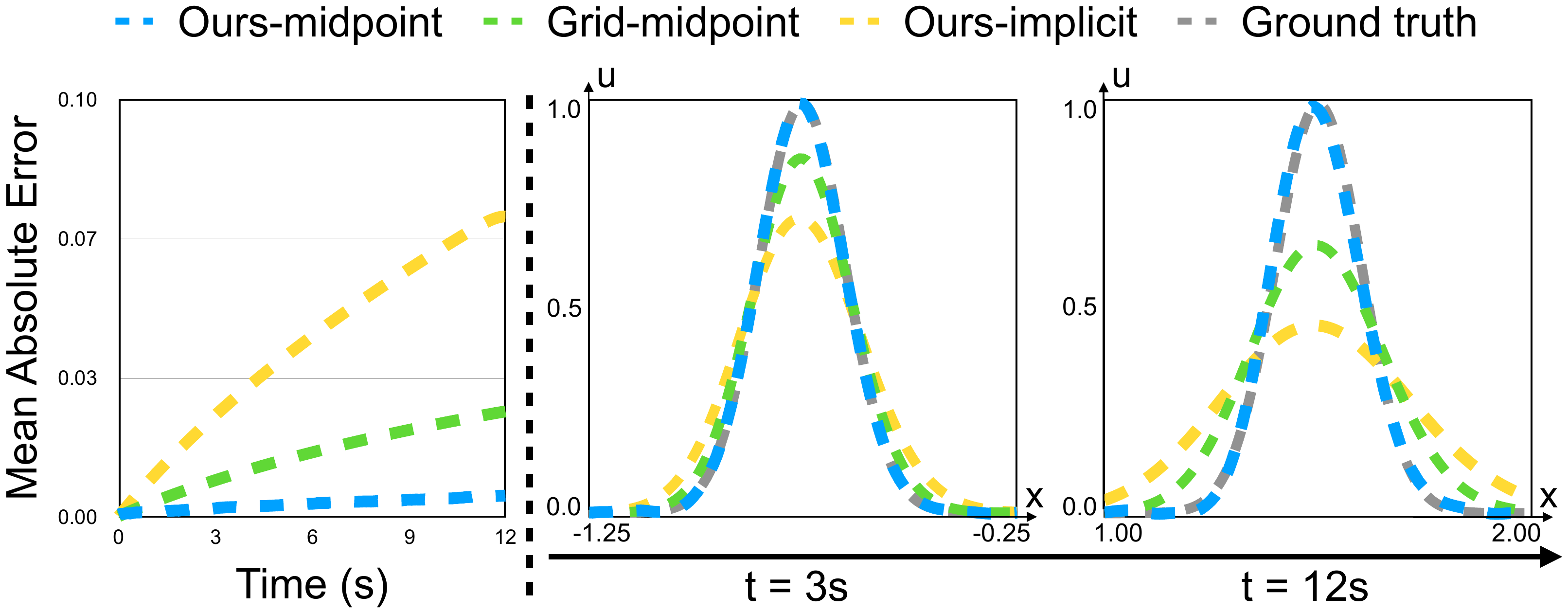}
  \end{center}
  \figspace
  \caption{\textbf{1D advection example:} 
    A Gaussian-shaped wave initially centered at $x=-1.5$ moves rightward with a constant velocity of $0.25$.
    From left to right, we show the mean absolute error plot over time and solutions at $t=3s$ and $t=12s$.
    Error is computed using $500$ uniform spatial points.
    Using an energy-preserving midpoint time integrator, our solution (blue) well approximates the ground truth (grey) over time,
    while the grid-based finite difference method (green) tends to diffuse over time. 
  }
  \vspace{-10pt}
  \label{advect_results}
\end{figure}

\begin{table}[t]
\centering
\caption{\textbf{Quantitative results for the 1D advection example (\Cref{advect_results}).} Error: mean absolute error over a total of $240$ time steps, compared to the ground truth analytical solution. Error is evaluated over $500$ uniform spatial samples. Time: runtime for a total of $240$ time steps. Memory: memory usage for storing the spatial representations.}
\label{advect_quant}
\begin{tabular}{lccc}
\toprule
Methods & Error    & Time  & Memory  \\
\midrule
Ours    & \textbf{0.0030} & 5.33h & \textbf{3.520KB} \\ 
Grid (same memory)    & 0.0146 & 1.13s & 3.520KB \\
Grid (same error)   & 0.0029 & 1.80s & 27.35KB \\ 	
\bottomrule
\end{tabular}
\end{table}

\paragraph{Baselines}
From classical solvers, we compare with three baselines: (1) the grid-based finite difference method \citep{fedkiw2001visual}, (2) the tetrahedral-mesh-based finite element method \citep{hughes2012finite}, (3) the meshless-particle-based material point method \citep{jiang2016mpm}. From neural-network-based, physics-informed approaches, we compare with another three baselines: (4) the original PINN \citep{raissi2019physics}, (5) PINN with temporal sub-domains \citep{krishnapriyan2021characterizing}, (6) physics-informed DeepONet \citep{wang2021learning,wang2021long}. We focus on comparing these physics-informed neural approaches because, like our method, they do not require any training data from the classic solvers.

\vspace{-10pt}
\paragraph{Comparison}
To strike an apple-to-apple comparison, we use the same time integrator and contact model for our approach and all the classical baselines. The only difference is the spatial representation. Notably, one can also use more advanced time integrators and contact models than the ones used in this work. Nevertheless, since the baselines and our approach adopt the same time integrators, our advantages on spatial discretization remain. For both classic and neural baselines, we ensure that they use the same amount of memory for storing the spatial representation as our method, e.g., grid size and the number of network layers.

We refer readers to \Cref{app:details,app:add_results} for other implementation details (e.g., initial / boundary conditions, baseline setups) and additional results. The temporal evolutions of the PDEs are best illustrated by the \textbf{supplementary video}.

\vspace{-10pt}
\subsection{Advection Equation}

Consider the classic 1D advection equation, 
\begin{align}
    \pd{\advectQuantity}{t} + (\advectSpeed\cdot\grad)\advectQuantity=0 \ ,
\end{align}
where $\advectSpeed$ is the advection velocity, and the vector field of interest is the advected quantity $\continuousField=\advectQuantity$.

\paragraph{Time Integration}
We adopt the same time integration scheme in both the discrete grid representation and ours. Choosing the energy-preserving midpoint method \citep{mullen2009energy} yields the time integration operator,
\begin{scriptsize}
\begin{align}
    \TimeIntegration &= \squaredNorm{\frac{\advectQuantityDiscretePlus(\pos)-\advectQuantityDiscrete(\pos)}{\dt} + (\advectSpeed\cdot\grad)(\frac{\advectQuantityDiscretePlus(\pos)+\advectQuantityDiscrete(\pos)}{2})} \ .
\end{align}
\end{scriptsize}

\paragraph{Results}
\Cref{advect_results} shows an example where a Gaussian-shaped wave moves with constant velocity $\advectSpeed=0.25$.
Under the same memory usage for storing the spatial representations, 
our approach uses $\mlpdepth=2$ hidden layers of width $\mlpwidth=20$,
and the finite difference grid resolution is $901$.

Using midpoint time integrator, the grid-based method (\texttt{grid-midpoint}) diffuses over time due to its spatial discretization, which is a well-known numerical issue \citep{courant1952solution,selle2008unconditionally}.
On the contrary, our result (\texttt{ours-midpoint}) does not suffer from numerical dissipation and agrees well with the ground truth at all frames.
We also tried the implicit Euler time integrator (see \texttt{ours-implicit}) and found it inherits its property of energy dissipation. Choosing the midpoint time integrator helps us preserve energy and obtain high-accuracy results.  In \Cref{advect_quant}, we report the quantitative evaluation result for our 1D advection example in \Cref{advect_results}.

 \begin{figure}[t]
   \begin{center}
   \includegraphics[width=\linewidth]{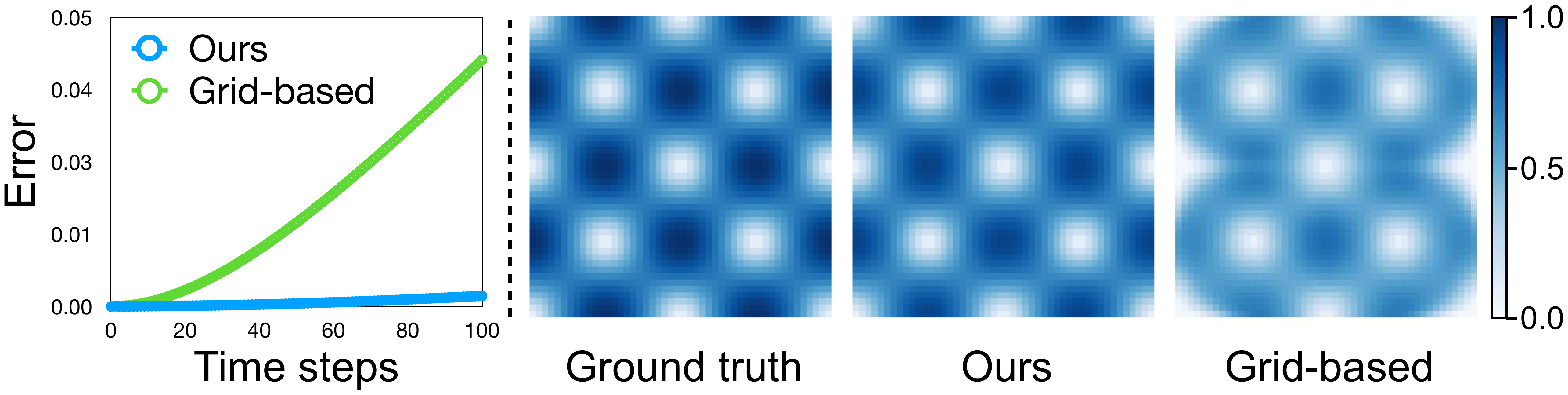}
   \end{center}
   \figspace
   \caption{\textbf{2D Taylor-Green vortex simulation.}
   	Left: the mean squared error of the velocity field for $100$ timesteps. Error is computed using $48^2$ uniform spatial points.
   	Right: velocity magnitude of solutions from the ground truth, ours, and the grid-based method (grid size $48$) at timestep $n=100$.
   	Under the same memory usage (for storing the spatial representation), our solution has a significantly smaller error than the grid-based method.
   }
   \label{taylorgreen_results}
 \end{figure}

\begin{table}[t]
\centering
\caption{\textbf{Quantitative results for the 2D Taylor-Green fluid example (\Cref{taylorgreen_results}).} Error: mean squared error of velocity field over total $100$ time steps, compared to the ground truth analytical solution. Time: runtime for a total of $100$ time steps. Memory: memory usage for storing the spatial representations.}
\label{taylorgreen_quant}
\begin{tabular}{lccc}
\toprule
Methods           & Error    & Time   & Memory \\
\midrule
Ours              & \textbf{3.35e-4} & 14.02h & \textbf{25.887KB}    \\
Grid (same memory) & 4.83e-3 & 2.91s  & 27.00KB    \\
Grid (same error) & 3.24e-4 & 189.4s  & 12.00MB    \\
\bottomrule
\end{tabular}
\end{table}

\vspace{-10pt}
\subsection{Incompressible Euler Equations}

 \begin{figure*}[t]
   \begin{center}
   \includegraphics[width=0.92\linewidth]{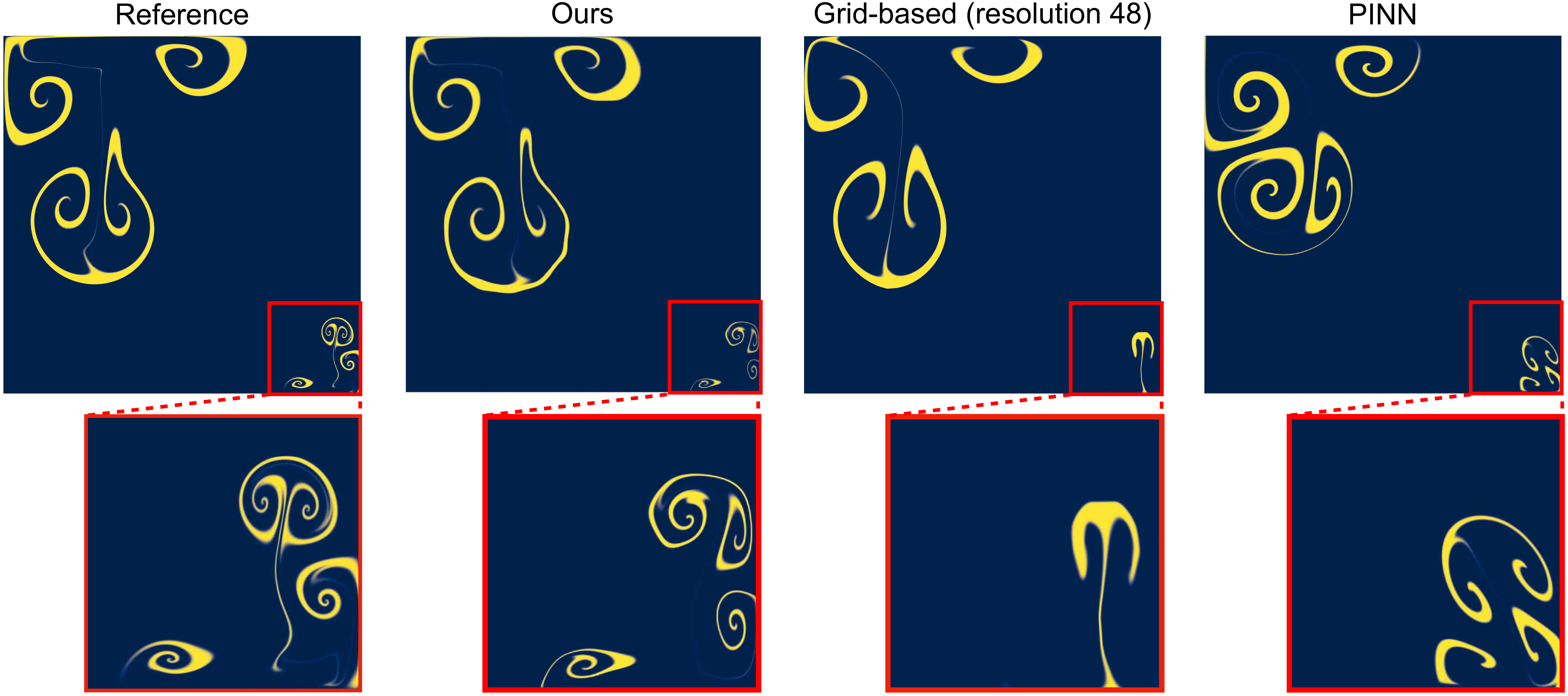}
   \end{center}
   \figspace
   \caption{\textbf{Two vortices of different scales.}
   We show the advected density field after $2.5$ seconds from the reference (top-left), our method (top-right), the grid-based method of resolution $48$ (bottom-left), and PINN (bottom-right). 
   The reference is obtained using the high-resolution grid-based method (we use resolution $1024$)
   and serves as a good approximation of the ground truth.
   Our MLP ($\mlpdepth=3,\mlpwidth=32$) has the same memory footprint as grids of resolution $48$.
   PINN uses the same MLP network as ours.
   Under the same memory constraint, our approach suffers less dissipation, captures more vorticity, and best resembles the reference solution, whose grids take $\sim 450\times$ memory compared to our network.
   See \Cref{fluid_adaptivity_init} for the initial condition of this example.
   }
   \figspace
   \label{fluid_adaptivity}
 \end{figure*}

 \begin{table}[t]
\figspace
\caption{\textbf{Quantitative results for the two-vortices fluid example in \Cref{fluid_adaptivity}.} 
Error: average absolute error of kinetic energy over a total of $50$ timesteps, compared to the reference solution. Kinetic energy is computed using $1024^2$ uniform samples. 
Time: runtime for a total of $50$ timesteps. 
Memory: memory usage for storing spatial representations. \texttt{Ours} and \texttt{Grid-based} \citep{stam1999stable} use the operator splitting scheme; \texttt{Ours-residual}, \texttt{PINN} \citep{raissi2019physics}, \texttt{PINN-sub} \citep{krishnapriyan2021characterizing} and \texttt{piDeepONet} \citep{wang2021learning} use the residual of the Euler equation as the objectives (see \Cref{pinn_objective} and \Cref{pinn_objective_discrete}).
}
\label{fluid_quant}
\centering

\begin{tabular}{lccc}
\toprule
Methods   & Error        & Time   & Memory \\
\midrule
Ours      & \textbf{2.24e2}  & 10.81h & 25.887KB    \\
Ours-residual & 2.97e4  & 10.07h & 25.887KB    \\
Grid-based & 1.07e4 & 1.78s  & 27.00KB     \\
PINN      & 2.25e4 & 7.88h   & 26.137KB    \\
PINN-sub & 3.21e4 & 20.83h & 26.137KB    \\
piDeepONet & 2.11e4 & 9.42h & 65.855KB    \\
\bottomrule
\end{tabular}
\figspace

\end{table}

In the incompressible Euler Equations
\begin{align}
\begin{split}
    \fluidDen(\pd{\fluidVelo}{t}+\fluidVelo\cdot\grad\fluidVelo) &= -\grad\press + \fluidDen\externalF,\\
    \grad\cdot\fluidVelo&=0,
\end{split}
\end{align}
the vector field of interest is the fluid velocity field $\continuousField=\fluidVelo$;
 $\press$ is the pressure, $\externalF$ is the external force, and $\fluidDen$ is the fluid density.
In our experiments, we consider $\fluidDen=1$ and $\externalF=0$.
The pressure field $\press$ is represented with another MLP network.

\vspace{-5pt}
\paragraph{Time Integration}
We apply the Chorin-style operator splitting scheme \citep{chorin1968numerical,stam1999stable} to both the neural spatial and finite-difference grid representations. This scheme converts the highly nonlinear PDE into three linear PDEs, which significantly eases the challenge of solving. The entire scheme breaks down into three sequential steps: advection (adv), pressure projection (pro), and velocity correction (cor).

\textbf{Advection} uses a semi-Lagrangian method, encoded by the operator~\citep{staniforth1991semi}
\begin{align}
    \TimeIntegrationAdv&=\squaredNorm{\fluidVeloAdvPlus(\pos)-\fluidVeloDiscrete(\posBT)} \ ,
\end{align}
whose optimization yields the advected velocity $\fluidVeloAdvPlus$. The backtracked location is given by $\posBT=\pos-\dt\fluidVeloDiscrete(\pos)$.
While traditional discrete representations compute the backtracked velocity using interpolation (e.g., linear basis function), our approach \emph{requires no interpolation}, only direct evaluation via network inference at $\posBT$.

\textbf{Pressure projection} is encapsulated by the operator
\begin{align}
    \TimeIntegrationPro&=\squaredNorm{\grad^2\pressPlus(\pos)-\grad\cdot\fluidVeloAdvPlus(\pos)}.
\end{align}
Plugging $\TimeIntegrationPro$ into the optimization solver, we obtain the pressure $\pressPlus$ that enforces incompressibility.
Note that the MLP that represents the velocity field $\fluidVeloAdv$ is kept fixed in this step.

\textbf{Velocity correction} is formulated by the operator
\begin{align}
    \TimeIntegrationCor&=\squaredNorm{\fluidVeloPlus-(\fluidVeloAdvPlus(\pos)-\grad\pressPlus(\pos))} \ , 
\end{align}
which adds the pressure gradient to the advected velocity yielding the \emph{incompressible} velocity $\fluidVeloPlus$.

\vspace{-5pt}
\paragraph{Results}
 
We first test our method on the 2D Taylor-Green vortex with zero viscosity \citep{taylor1937mechanism,brachet1983small}. The closed-form analytical solution is given by: $\fluidVelo(\pos, t)=(\sin x \cos y, -\cos x \sin y)$ for $\pos \in [0,2\pi]\times[0,2\pi]$.
To compare under the same memory usage (for storing the velocity field), we use $\mlpdepth=3$ hidden layers of width $\mlpwidth=32$ for our MLP and set the grid size to $48$ for the grid-based projection method \citep{stam1999stable}.
We set $\dt=0.05$ and execute both methods for $100$ timesteps.
In \Cref{taylorgreen_results}, we show the mean squared error of the solved velocity field over time. In \Cref{taylorgreen_quant}, we report the quantitative evaluation result for our 2D Taylor-Green example in \Cref{taylorgreen_results}.
This example demonstrates that our method excellently preserves a stationary solution and obtains a much smaller error than the grid-based method.

For discrete grid representation, efficiently capturing multi-scale details usually requires difficult-to-implement adaptive data structures \citep{setaluri2014spgrid}. 
Instead, INSRs are adaptive by construction \citep{xie2021neural} and enable us to capture more details under the same memory storage.
We set up an example where the initial velocity field is composed of two Taylor-Green vortices of different scales (see \Cref{fluid_adaptivity_init} for illustration).

\begin{figure}[t]
    \begin{center}
    \includegraphics[width=\linewidth]{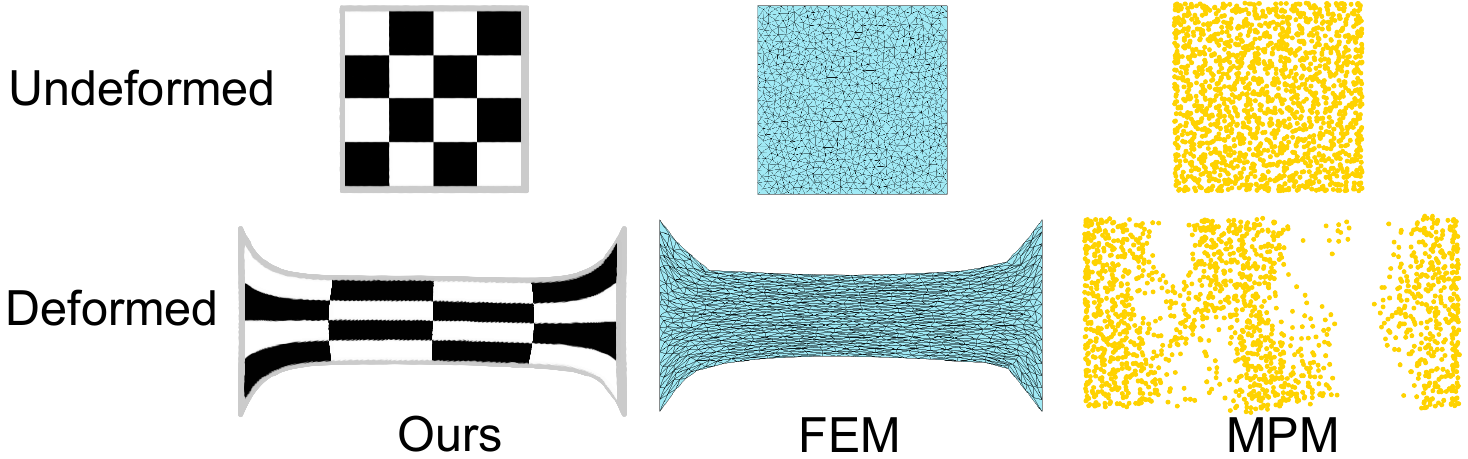}
    \end{center}
    \figspace
    \caption{\textbf{Elastic tension test}. 
    We use $\alpha = 3$ hidden layers of width $\beta = 68$ for our MLP, which takes the same memory as the FEM mesh (0.8K vertices, 1.5K faces) and MPM point cloud (1.7K points).
    }
    \label{2d_patch_test_vs_fem_mpm}
    \vspace{-5pt}
\end{figure}

\begin{figure}[t]
  \begin{center}
  \includegraphics[width=\linewidth]{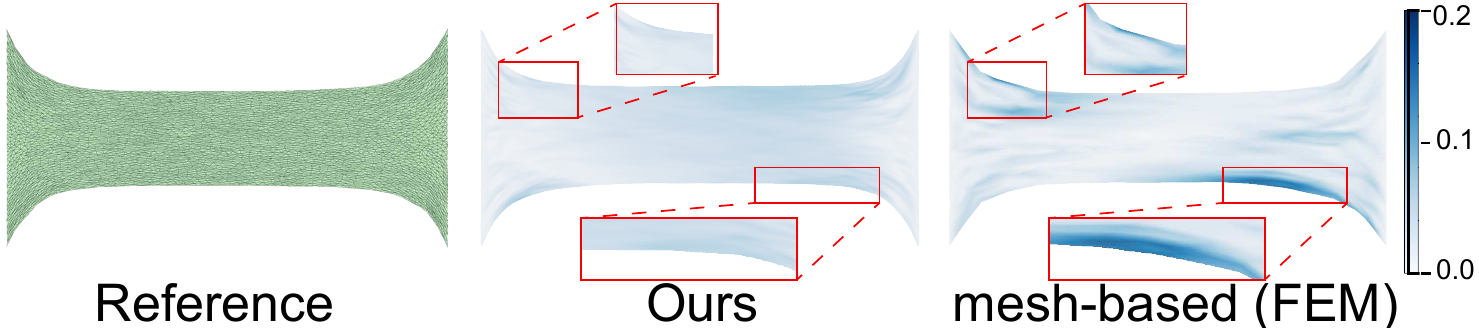}
  \end{center}
   \figspace
  \caption{\textbf{Error of the elastic tension test.}
  We visualize the L2 distance error of our result and the FEM result (0.8K vertices). Errors are calculated against the reference result obtained by high-resolution FEM. Our result obtains smaller errors.
  }
\figspace
  \label{2d_patch_test_error}
\end{figure}

In \Cref{fluid_adaptivity} and \Cref{fluid_quant}, we show our results on this example and compare with the grid-based projection method \citep{stam1999stable}, PINN \citep{raissi2019physics}, PINN with temporal sub-domains (PINN-sub) \citep{krishnapriyan2021characterizing} and piDeepONet \citep{wang2021learning}.
All methods are compared under the same memory storage for spatial representations.
For PINN and PINN-sub, we use the same MLP structure as ours.
Detailed setups for these baselines can be found in \Cref{fluid_appendix}.
We execute our approach, the grid-based method, and PINN-sub for $50$ timesteps with $\dt=0.05$, and train PINN and piDeepONet with the same temporal range of $2.5$ seconds.
Our approach can capture the fine details of the smaller vortex, best approximate the reference solution, and has the most negligible energy dissipation.

\begin{table}[t]
    \centering
    \caption{\textbf{Quantitative results for the 2D elasticity tension example in \Cref{2d_patch_test_error}.}
    Error: infinity norm of L2 distance w.r.t. the high-resolution ground truth. Time: total runtime until convergence. Memory: memory usage for storing spatial representations.
    }
    \label{tension_quant}
    \begin{tabular}{lccc}
    \toprule
    Methods   & Error        & Time   & Memory \\
    \midrule
    Ours      & \textbf{8.82e-2} & 38.33m & 56.32KB    \\
    Mesh-based & 1.99e-1 & 22.04s & 54.00KB   \\
    \bottomrule
    \end{tabular}
\figspace
\end{table}

Moreover, the superior accuracy of our approach also comes from the usage of the operator-splitting time integration scheme.
While PINN, PINN-sub, and piDeepONet also use implicit neural representations like ours, they treat time as a continuous variable (i.e., part of the network inputs).
Therefore, they cannot use the operator-splitting scheme but only employ the residual of the Euler equations as the training objective, which is more challenging to optimize. The time-discrete PINN proposed by \citet{raissi2019physics} specializes in Runge-Kutta schemes and does not support operator splitting in its current form either. We verify these observations by changing the objectives of our method to a similar training objective used by them, i.e., the residual of the Euler equation (\Cref{pinn_objective}) with a fully-implicit-time-discretization (\Cref{pinn_objective_discrete}). After such change, we obtain a significantly worse result (see \texttt{ours-residual} in \Cref{fluid_quant}), which confirms the necessity of using the operator-splitting scheme. This experiment demonstrates that just replacing PINN/SIREN's time-dependent formulation is insufficient, but the particular choice of temporal discretization matters. 

\begin{figure*}[t]
    \begin{center}
    \includegraphics[width=\linewidth]{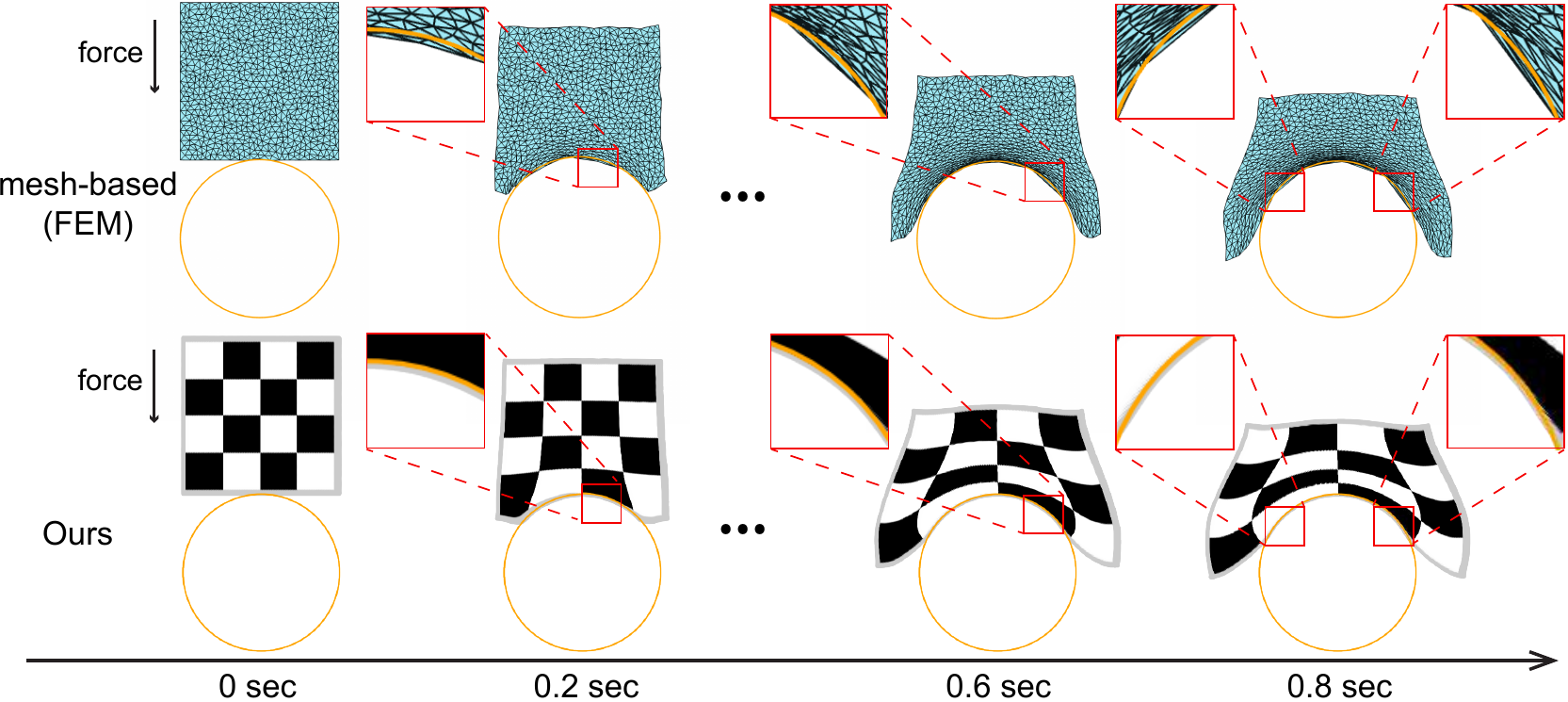}
    \end{center}
    \figspace
    \caption{\textbf{An elastic square collides with a circle.} 
    Under the same memory footprint (for storing the spatial representation), the result of mesh-based FEM (top) conforms poorly at the collision interface. In contrast, our result (bottom) fits the boundary more gracefully.
    }
    \figspace
    \label{2d_square_collide_sphere}
\end{figure*}

\vspace{-10pt}
\subsection{Elastodynamic Equation}
\label{sec:elasticity}
Lastly, we study the Elastodynamic equation 
\begin{align}
    \solidDen\ddot{\deformMap} = \grad \cdot \pkstress(\deformGrad) + \solidDen\bodyforce 
\end{align}
that describe the motions of deformable solids \citep{gonzalez2008first}. 
The vector field of interest is the deformation map $\continuousField=\deformMap$. 
Here $\solidDen$ is the density in the reference space, $\pkstress$ is the first Piola-Kirchhoff stress, $\deformGrad=\grad\deformMap$ is the deformation gradient, $\dot{\deformMap}$ and $\ddot{\deformMap}$ are the velocity and acceleration, and $\bodyforce$ is the body force.

We assume a hyper-elasticity constitutive law, i.e., $\pkstress=\pdflat{\energyDensity}{\deformGrad}$, where $\energyDensity$ is the energy density function. In particular, we assume a variant of the stable Neo-Hookean energy \citep{smith2018stableNH}
\begin{align}
    \energyDensity = \frac{\firstlame}{2}\trace^2(\singValues-\identity) +  \secondlame(\determinant(\deformGrad)-1)^2,
\end{align}
where $\firstlame$ and $\secondlame$ are the first and second lame parameters, $\singValues$ are the singular values of the deformation gradient $\deformGrad$, and $\determinant(\deformGrad)$ is the determinant of the deformation gradient $\deformGrad$. When $\secondlame=0$, the elastic energy recovers the As-Rigid-As-Possible energy \citep{sorkine2007rigid}.

\vspace{-10pt}
\paragraph{Time Integration}
We apply the variational time integration scheme \citep{gast2015optimization,kane2000ip} to the (1) tetrahedral finite element method, (2) the material point method, and (3) our neural representation,
\begin{align}
\begin{split}
    \TimeIntegration = &\underbrace{\frac{1}{2}\solidDen(\defomMapDotDiscretePlus-\defomMapDotDiscrete)^T(\defomMapDotDiscretePlus-\defomMapDotDiscrete)}_{\text{kinetic energy}} + \\
    &\underbrace{\vphantom{\frac{1}{2}}\energyDensity(\defomMapDiscretePlus)}_{\substack{\text{elastic} \\ \text{energy}}} \underbrace{\vphantom{\frac{1}{2}} - \solidDen\bodyforce^T\defomMapDiscretePlus}_{\substack{\text{external force} \\ \text{potential}}} \ ,
    \label{eqn:int:solid}
\end{split}
\end{align}
where $\defomMapDotDiscretePlus=(\defomMapDiscretePlus-\defomMapDiscrete)/ \dt$, $\solidDen$ is the density, $\bodyforce$ is the external force. We can also incorporate boundary conditions, e.g., positional and contact constraints, by introducing additional energy terms \citep{bouaziz2014projective,li2020incremental} (see \Cref{app:bc_elasticity}). These energy terms allow us to simulate challenging contact problems where the material impacts a collision surface at high speed.

\vspace{-5pt}
\paragraph{Results} 

\begin{table}[t]
    \centering
    \vspace{-8pt}
    \caption{\textbf{Quantitative results for the collision example in \Cref{2d_square_collide_sphere}.} Error: maximum overlapping distance between the square and the circle. Time: runtime for a total of $10$ timesteps. Memory: memory usage for storing spatial representations.
    }
    \label{collide_sphere_quant}
    \resizebox{\linewidth}{!}{
    \begin{tabular}{lccccc}
      \toprule
       & \multicolumn{3}{c}{ Error } & \multirow{2}{*}{ Time } & \multirow{2}{*}{ Memory} \\
      Methods & 0.2s & 0.6s  & 0.8s & & \\
      \midrule
       Ours & 1.62e-2 & 1.04e-2 & 1.06e-2 & 23.0m & 56.32KB \\
       Mesh-based & 3.45e-2 & 5.47e-2 & 4.23e-2 & 98.2s & 54.00KB \\
      \bottomrule
    \end{tabular}
    }
    \figspace
    \vspace{-2pt}
\end{table}

We first evaluate our method on a typical 2D example for elastic tension test, 
and compare with the traditional finite element method (FEM, \citep{hughes2012finite,reddy2019fem}) using tetrahedral mesh representation. 
We use $\alpha = 3$ hidden layers of width $\beta = 68$ for our MLP, which takes the same memory as the meshes used by FEM (0.8K vertices, 1.5K faces). 
The geometry is rendered as point clouds for our method in all our examples.
As shown in \Cref{2d_patch_test_error} and \Cref{tension_quant}, our method obtains a more minor error than FEM.
Classic mesh-less particle technique (material point method, MPM) suffers incorrect numerical fracture in this example (see \Cref{2d_patch_test_vs_fem_mpm}).

\begin{figure}[t]
    \begin{center}
    \includegraphics[width=\linewidth]{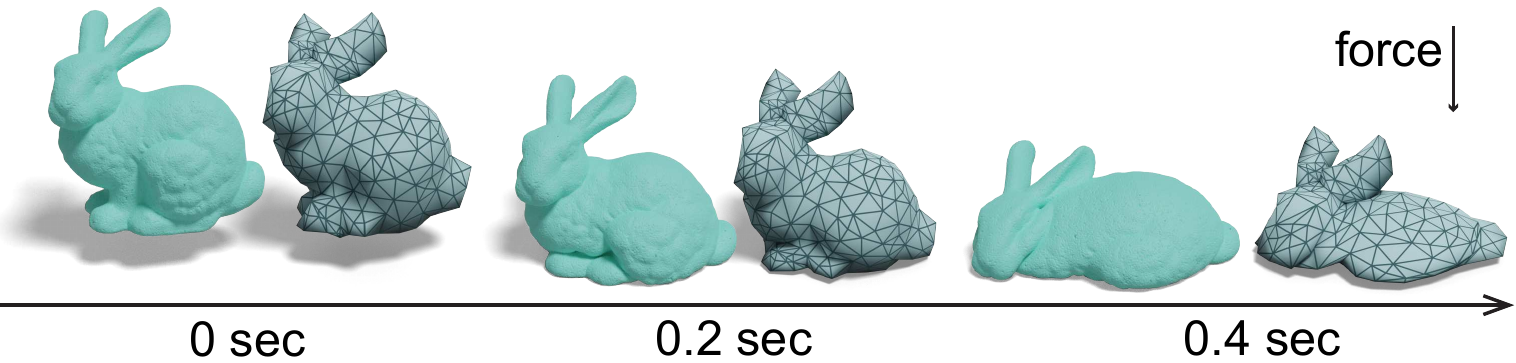}
    \end{center}
    \figspace
    \vspace{-5pt}

    \caption{\textbf{A bunny collides with the ground in 3D. } Using INSR, our method (green, left) captures more intricate geometry details and complex dynamics compared to the traditional mesh-based FEM (blue, right) under the same memory usage.
    }
    \figspace
    \label{3d_bunny_collide_plane}
\end{figure}

\begin{figure}[t]
  \begin{center}
    \includegraphics[width=\linewidth]{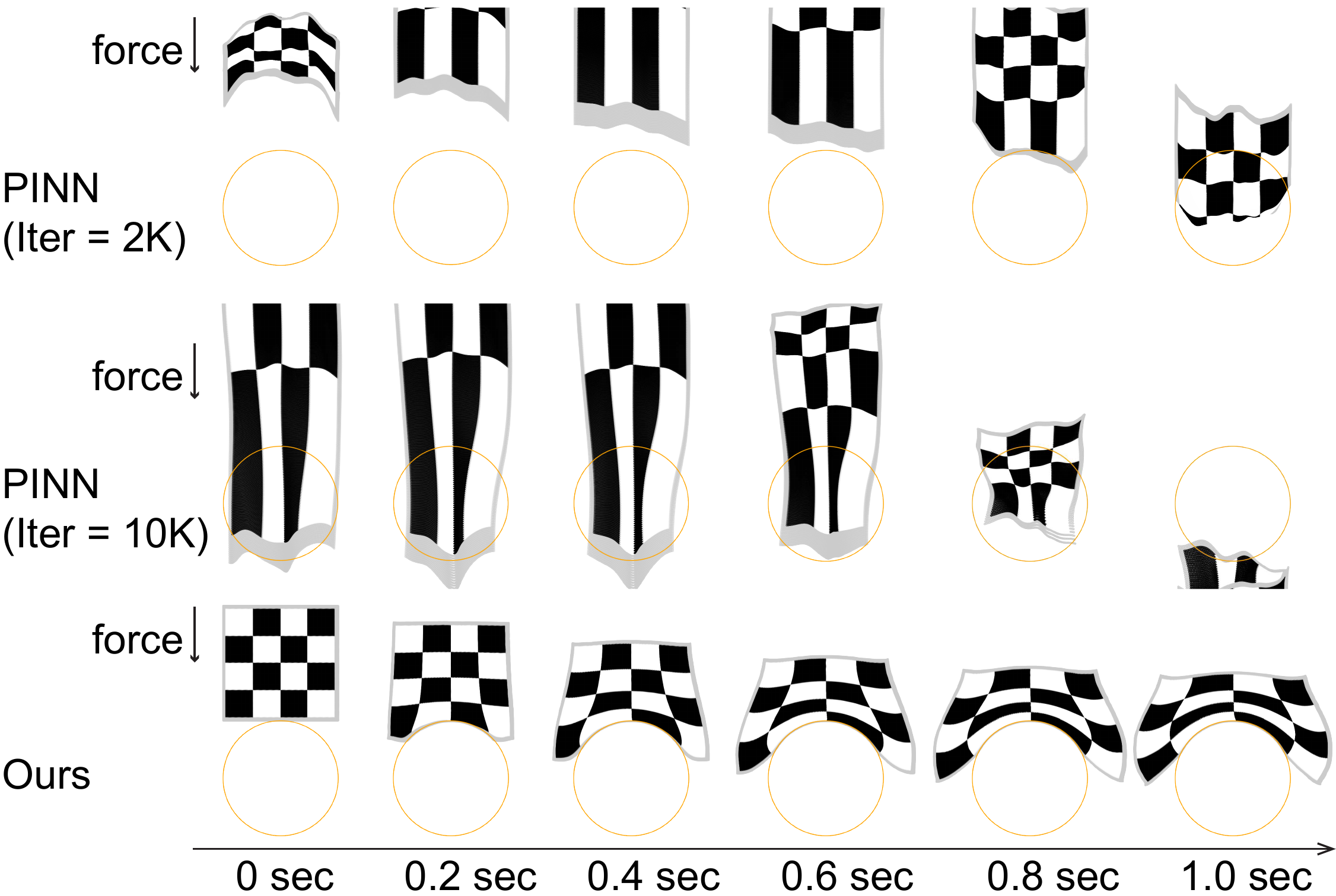}
    \includegraphics[width=\linewidth]{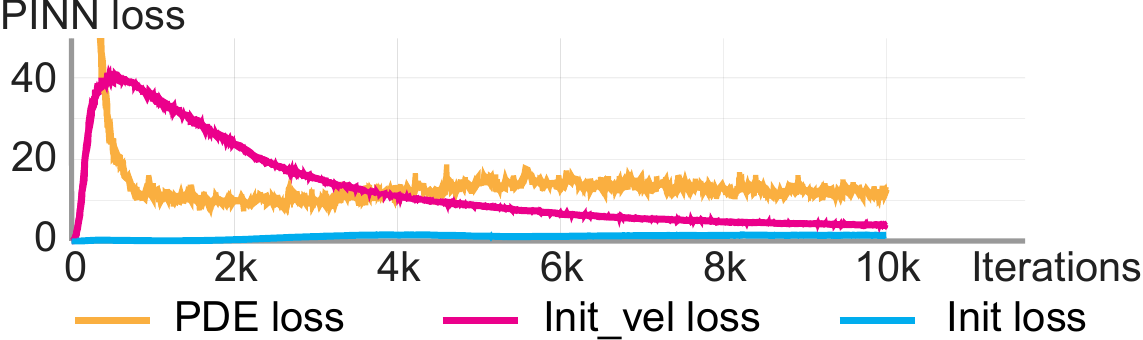}
     \end{center}
    \figspace
        \caption{
    \textbf{PINN's result on the square-circle collision example in \Cref{2d_square_collide_sphere}.} Our approach successfully captures the correct contact behaviors, but PINN fails to do so under the same memory usage and network structure. To demonstrate this, we present PINN's failure results trained with both 2K and 10K ADAM iterations (top).
     Despite making progress in minimizing the loss (bottom) under various hyper-parameters, such as learning rate and collision ratio, PINN was unable to properly capture the highly discontinuous contact elasticity. Moreover, even with additional training iterations (i.e., more than 10K), the training loss did not improve.
        }
      \label{pinn_comparison}
       \vspace{-18pt}
\end{figure}

By using INSR, our method can capture more intricate details than the traditional discrete representations under the same memory usage.
In \Cref{2d_square_collide_sphere}, we show that our method allows the deformed square to gracefully fit the boundary of the sphere during the non-trivial collision. 
In contrast, the mesh-based FEM struggles to produce smooth results due to its insufficient mesh resolution. 
To alleviate such artifacts, the mesh-based FEM either needs to increase resolutions, thus inducing higher memory cost, or conducts complex remeshing \citep{narain2012adaptive}. 
As shown in \Cref{3d_bunny_collide_plane} and \Cref{3d_lucy_collide_plane}, our method allows for more complex dynamics and fine geometry details compared to the mesh-based FEM under the same memory footprint.

Note that we adopt the same collision detection and handling strategy for both the neural representation and the mesh-based representation (FEM). 
Specifically, we use a spring-like penalty force and the corresponding energy to move the collided point out of its collision surface, similar to \citep{mcadams2011efficient, xian2019multigrid}. 
Since our approach and FEM share the same time integration scheme and the same collision handling method, the difference strictly stems from the underlying spatial representations.

Finally, in \Cref{pinn_comparison}, we compare our time-independent formulation to PINN's time-dependent approach on non-trivial collision cases. PINN struggles to capture the correct contact behavior when extreme nonlinearities and discontinuities involve.
The most likely reason is that PINN models the time continuously, but collision is highly discontinuous in time. Therefore, the loss function of PINN is prone to abrupt change when the collision happens, and collision forces are involved. Our approach is able to adopt the variational time integrator and use the incremental potential as our loss function, which is known for its stability.

\vspace{-10pt}
\section{Discussion and Conclusion}
This work explores INSR for numerically modeling time-dependent PDEs. Combined with a wide range of classical time integrators, the INSR solver captures various advection, elasticity, and fluid phenomena and outperforms previous physics-informed neural network approaches on multiscale, turbulent flows and contact mechanics problems. Compared to classic \emph{explicit} representations (e.g., grids and meshes), our approach offers improved accuracy, reduced memory, and automatic adaptivity.

While offering important benefits, INSR-based PDE time-stepping requires longer wall-clock computation time than existing methods. (For reference, PINN also takes longer to solve forward problems than classic FEMs. See also Table 1 by \citet{zehnder2021ntopo} and Section 7 by \citet{yang2021geometry}.) Optimizing \emph{globally-supported} neural network weights takes longer than optimizing \emph{locally-supported} grid values, even if there are fewer neural network weights than the number of grid nodes. For instance, for the bunny example (\Cref{3d_bunny_collide_plane}), our neural network optimization takes around 30 minutes per timestep while the corresponding FEM simulation takes less than 1 minute. 

Facing this wall clock vs. memory/accuracy/adaptivity trade-off, we believe an exciting future direction is hybrid mesh-neural spatial representations that aim for the best of both worlds. Indeed, our work does not advocate INSR as the ``perfect'' spatial representation for solving PDEs. Instead, we view our work as a stepping stone toward future hybrid mesh-neural PDE solvers. For this matter, recent hybrid representation works \citep{muller2022instant,takikawa2021neural,martel2021acorn} have offered promising results in reducing training time from hours to seconds while keeping the expressiveness of INSR. We hope our work will serve as a benchmark for future representations.

Our work demonstrates the effectiveness of INSR in solving time-dependent PDEs and observes empirical convergence under refinement (see \Cref{3d_elasticity_convergence}). Future work may consider a theoretical analysis of convergence and stability. 
More challenging physical phenomena, such as turbulence and intricate contacts, are also important future directions. 
Currently, our work enforces ``soft'' boundary conditions. Enforcing ``hard'' boundary conditions on a neural network is another exciting direction \citep{lu2021physics}.

\vspace{-10pt}
\section*{Acknowledgements}
This research was partially supported by National Science Foundation (1910839) and a Natural Sciences and Engineering Research Council of Canada (NSERC) Discovery Grant [RGPIN-2021-03733]. We thank Henrique Teles Maia for proofreading and all the anonymous reviewers for their helpful comments and suggestions. Lastly, we want to thank all our math professors (James Scott, Joseph Teran, among many others) who instilled in us a love for PDEs.


\bibliography{main}
\bibliographystyle{icml2023}

\clearpage
\appendix
\newpage




\section{Implementation Details}
\label{app:details}

\subsection{Optimization}
We solve our time-integration optimization problem (\Cref{eqn:opti-time-inte-nn}) with the Adam optimizer \citep{kingma2014adam}.
For all examples in our experiments, we set an initial learning rate $\lrInit$ and reduce it by a factor of $0.1$ if the loss value does not decrease for $\iterPatience$ iterations.
We stop the optimization process when the learning rate is lower than $\lrMin$ or until it reaches a maximum of $\iterMax$ iterations.
Specific values of these hyper-parameters are described for each example below.
We implemented our method using the PyTorch library and performed experiments on an NVIDIA GeForce RTX 3090 GPU.


\subsection{Advection Equation}
For our advection example in \Cref{advect_results}, the 1D spatial domain is $\spatialDomain=[-2,2]$.
We consider the Dirichlet boundary condition, i.e., the advected quantity at boundaries equals zero. 
Hence we set the boundary constraint term in \Cref{eqn:opti-time-inte-nn-bc} as 
\begin{align}
	\BoundaryConstraint=||\advectQuantityDiscretePlus(\pos)||_2^2,
\end{align}
with the weighting factor $\lambda=1$. 
The initial condition for this example is 
\begin{align}
	\hat{\advectQuantity^0}(\pos) = e{-\frac{(\pos-\mu)^2}{2\sigma^2}},
\end{align}
with $\mu=-1.5$ and $\sigma=0.1$. 
We set the optimization hyper-parameters $\lrInit=1e{-4}$, $\lrMin=1e{-8}$, $\iterPatience=500$ and $\iterMax=20000$.
For each gradient descent iteration, we randomly sample $|\sampleSet|=5000$ points within the spatial domain $[-2,2]$.
For this example, our method takes $\sim80s$ to compute per timestep, while the grid-based method (using the same memory) takes $\sim 4e{-3}s$.

\subsection{Incompressible Euler Equations}
\label{fluid_appendix}
For our 2D fluid examples, the spatial domain is $\spatialDomain=[-1, 1]\times[-1, 1]$. 
We consider solid boundary conditions, i.e., the fluid cannot go through the boundaries.
Recall that we adopt the operator splitting scheme.
Therefore, the boundary constraint terms for the three sequential steps are
\begin{align}
\begin{split}
	\BoundaryConstraint_{adv} = ||\fluidVelo_{adv\perp}^{n+1}(\pos)||_2^2 \\
	\BoundaryConstraint_{pro} = ||\grad_\perp \press^{n+1}(\pos)||_2^2 \\
	\BoundaryConstraint_{cor} = ||\fluidVelo_\perp^{n+1}(\pos)||_2^2 \\
\end{split}
\end{align}
where $\perp$ indicates the perpendicular direction against the boundary. The weighting factor $\lambda=1$.

\begin{figure}[t]
   \begin{center}
   \includegraphics[width=0.99\linewidth]{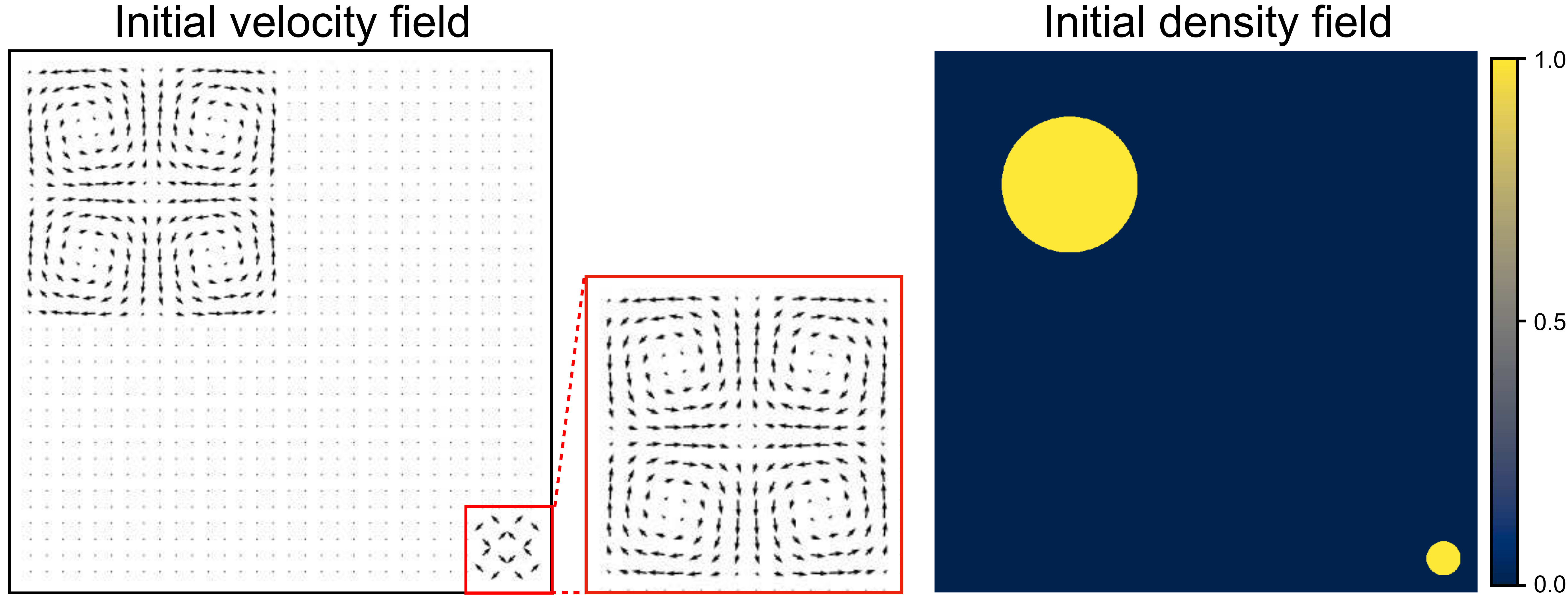}
   \end{center}
   \figspace
   \caption{\textbf{Initial condition for the example in \Cref{fluid_adaptivity}.}
   Left: velocity field (\Cref{Eqn:multi-vortices-velo}). Right: density field (\Cref{Eqn:multi-vortices-den}).
   }
   \label{fluid_adaptivity_init}
\end{figure}
 
\paragraph{2D Taylor-Green vortex}
Standard 2D Taylor-Green is originally defined in domain $[0, 2\pi]\times[0, 2\pi]$.
We translate and scale the domain to $[-1, 1]\times[-1,1]$ such that the input range fits our MLP with the SIREN activation \citep{sitzmann2020implicit}.
Therefore, the initial condition for the velocity field becomes
\begin{align}
\begin{split}
	\hat{\fluidVelo^0}(\pos)=(&\frac{1}{\pi}\sin[\pi(x+1)]\cos[\pi(y+1)], \\
 &-\frac{1}{\pi}\cos[\pi(x+1)]\sin[\pi[y+1]]).
\end{split}
\end{align}
After the simulation, we convert it back to domain $[0, 2\pi]\times[0, 2\pi]$ for evaluation and comparison.
We set the optimization hyper-parameters $\lrInit=1e{-5}$, $\lrMin=1e{-8}$, $\iterPatience=500$ and $\iterMax=20000$.
The size of the sample set is $|\sampleSet|=256^2$.
For this example, our method takes $\sim10\text{min}$ to compute per timestep, while the grid-based method (using the same memory) takes $\sim 0.03s$.

\begin{table*}[t]
  \setlength{\tabcolsep}{3pt}
  \caption{\textbf{Experiment setup for the elasticity examples.} $\solidDen$ is the density. $\firstlame$ and $\secondlame$ are the first and second lame parameters. $\alpha$ and $\beta$ are the number of hidden layers and the dimension of the hidden features. $\lrInit$ and $\iterMax$ are the initial learning rate and the maximum number of iterations. $\timeAvg$ is the average training time per time step. Note that the density $\solidDen$ and timestep size \textbf{dt} are reported as N/A for the quasistatic example Stretch (2D) (\Cref{2d_patch_test_vs_fem_mpm}).
  }
  \label{tbl:elasticity_setup}
  \begin{center}
  \begin{tabular}{l c c c c c c c c c c c}
   \textbf{Example} & \textbf{Dim} & $|\sampleSet|$ & \textbf{dt}  &   $\solidDen$ & $\firstlame$ & $\secondlame$ & $\alpha$ & $\beta$ & $\lrInit$ & $\iterMax$ & $\timeAvg$(s)\\
   \hline 
   \rowcolor[HTML]{DAE8FC} 
   \textbf{Collision (2D) (\Cref{2d_square_collide_sphere})}   & 2    & $100^2$  & 0.1 & $1 e1$ & $2 e1$ & $1 e3$ & 3 & 68 & $1e{-5}$ & $1 e{4}$ & $1.38e{2}$ \\
   \textbf{Stretch (2D) (\Cref{2d_patch_test_vs_fem_mpm})}   & 2    & $100^2$  & N/A & N/A & $1 e0$ & $1 e{3}$ & 3 & 68 & $1 e{-4}$ & $5 e{4}$ & $2.30e{3}$\\
   \rowcolor[HTML]{DAE8FC} 
   \textbf{Bunny (\Cref{3d_bunny_collide_plane})}         & 3     & $20^3$  & 0.1  &  $1 e0$ & $1 e2$ & $1 e3$ & 3 & 66 & $1 e{-5}$ & $2 e{4}$ & $1.70e{3}$\\
   \textbf{Spot (\Cref{fig:opti-time-inte-nn})}    & 3   & $20^3$ & 0.1  &  $1 e0$ & $1 e2$ & $1 e3$ & 3 & 66 & $1 e{-3}$ & $5 e{3}$ & $1.74e{3}$\\
   \rowcolor[HTML]{DAE8FC} 
   \textbf{Lucy (\Cref{3d_lucy_collide_plane})}   & 3  & $20^3$  & 0.1  & $1 e0$ & $1 e3$ & $1 e3$ & 3 & 128 & $1 e{-4}$ & $2 e{4}$ & $1.16e{3}$\\
   \hline
  \end{tabular}
  \end{center}
	\figspace  
  \end{table*}

\paragraph{Two vortices of different scale}

For the example shown in \Cref{fluid_adaptivity}, the initial condition for the velocity field is
\begin{align}
	\hat{\fluidVelo^0}(\pos)=
	\begin{cases}
		\hat{\fluidVelo^1}(\pos), & \pos \in [-1, 0]^2 \\
		\hat{\fluidVelo^2}(\pos), & \pos \in [\frac{7}{4}, 1]^2 \\
		(0, 0), & \text{otherwise.}	
	\end{cases}
	\label{Eqn:multi-vortices-velo}
\end{align}
where
\begin{align}
    \begin{split}
    \hat{\fluidVelo^1}(\pos) = &(\sin[2\pi(x + 1)] \cos [2\pi(y + 1)], \\
    &-\cos [2\pi(x + 1) ] \sin [2\pi(y + 1)]),
    \end{split} \\
    \begin{split}
    \hat{\fluidVelo^2}(\pos) = &(\sin[8\pi(x - \frac{7}{4})] \cos [8\pi(y - \frac{7}{4})], \\
    &-\cos [8\pi(x - \frac{7}{4}) ] \sin [8\pi(y - \frac{7}{4})]).
    \end{split}
\end{align}

The density field that we advect is initialized as 
\begin{align}
	\hat{d^0}(\pos)=
	\begin{cases}
		1 & ||2\pos + 1 || \le 0.5 \text{ or } ||8\pos + 7 || \le 0.5\\
		0 & \text{otherwise.}	
	\end{cases}
	\label{Eqn:multi-vortices-den}
\end{align}
\Cref{fluid_adaptivity_init} visually illustrates the above initial conditions.
We set the optimization hyper-parameters $\lrInit=1e{-5}$, $\lrMin=1e{-8}$, $\iterPatience=500$ and $\iterMax=20000$.
The size of the sample set is $|\sampleSet|=128^2$.
For this example, our method takes $\sim10\text{min}$ to compute per timestep.

\paragraph{Baseline setups}
For PINN \citep{raissi2019physics} and PINN-sub \citep{krishnapriyan2021characterizing}, we use the same MLP structure as ours ($\mlpdepth=3$ hidden layers of width $\mlpwidth=32$, with SIREN activation).
For piDeepONet \citep{wang2021learning}, we use $\mlpdepth=3$ hidden layers of width $\mlpwidth=32$ with Tanh activation for both the branch net and trunk net.
Since they do not support the operator-splitting scheme, we follow the previous literature \citep{chuang2022experience} and use the residual of incompressible Euler equation as the physics-informed training objective, 
\begin{small}
\begin{align}
\label{pinn_objective}
\begin{split}
\mathcal{L}_{\theta_\fluidVelo, \theta_\press} &= 
\underbrace{\sum_{\substack{\pos\in\sampleSet\subset\spatialDomain\\t\in\temporalDomain}} ||\pd{\fluidVelo}{t}+\fluidVelo\cdot\grad\fluidVelo + \grad\press||_2^2 + ||\grad\cdot\fluidVelo||_2^2}_\text{PDE residual} \\
&+ \underbrace{\sum_{\substack{\pos\in\spatialDomain\\t = 0}} ||\fluidVelo - \hat{\fluidVelo^0}||_2^2}_\text{initial condition}
+ \underbrace{\sum_{\substack{\pos\in\partial\spatialDomain\\t\in\temporalDomain}} ||\fluidVelo_\perp(\pos)||_2^2}_\text{boundary condition}
\end{split}
\end{align}
\end{small}
where $\fluidVelo=\fluidVelo_{\theta_\fluidVelo}(\pos, t)$ and $\press=\fluidVelo_{\theta_\press}(\pos, t)$ are parameterized by MLPs.
The number of spatial samples used for each training iteration is the same as ours.
We train their models until convergence.

Another baseline ($\texttt{Ours-residual}$ in \Cref{fluid_quant}) uses an implicit-time-discretized version of this objective function for time integration, i.e.,
\begin{small}
\begin{align}
\label{pinn_objective_discrete}
\begin{split}
\TimeIntegration &= 
\sum_{\substack{\pos\in\sampleSet\subset\spatialDomain\\t\in\temporalDomain}} ||\frac{\fluidVeloPlus - \fluidVeloDiscrete} {t}+\fluidVeloPlus\cdot\grad\fluidVeloPlus + \grad\pressPlus||_2^2 \\
&+ \sum_{\substack{\pos\in\sampleSet\subset\spatialDomain\\t\in\temporalDomain}} ||\grad\cdot\fluidVeloPlus||_2^2
+ \sum_{\substack{\pos\in\partial\spatialDomain\\t\in\temporalDomain}} ||\fluidVeloPlus_\perp(\pos)||_2^2.
\end{split}
\end{align}
\end{small}

\begin{figure*}[!thb]
    \centering
    \begin{minipage}{0.24\linewidth}
        \centering
        \includegraphics[width=\linewidth]{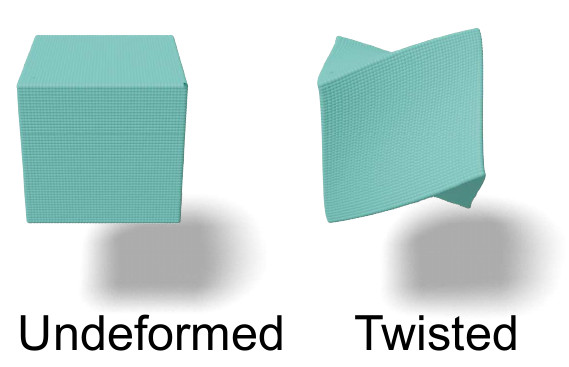}
        \figspace
        \caption{\textbf{Twisting test.} Quasistatic simulation in 3D. The right end is twisted 45 degrees.}
        \label{3d_twisting_test}
    \end{minipage}%
    \qquad
    \begin{minipage}{0.7\textwidth}
        \centering
        \includegraphics[width=\linewidth]{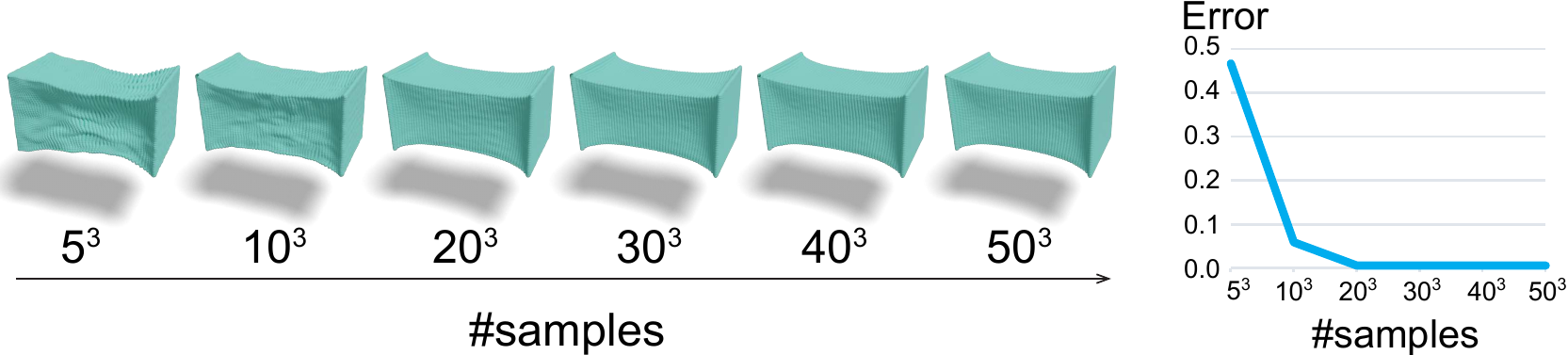}
       \figspace
      \caption{\textbf{Sampling convergence test}. We use a different number of spatial samples $|\sampleSet|$ and optimize for the same number of gradient descent iterations. On the right, we further report the error with respect to the reference result (using $|\sampleSet| = 50^3$). As we increase the number of spatial samples, the simulation result converges.
      }
      \label{3d_elasticity_convergence}
    \end{minipage}
    \vspace{-10pt}
\end{figure*}

\begin{figure*}[t]
  \begin{center}
  \includegraphics[width=\linewidth]{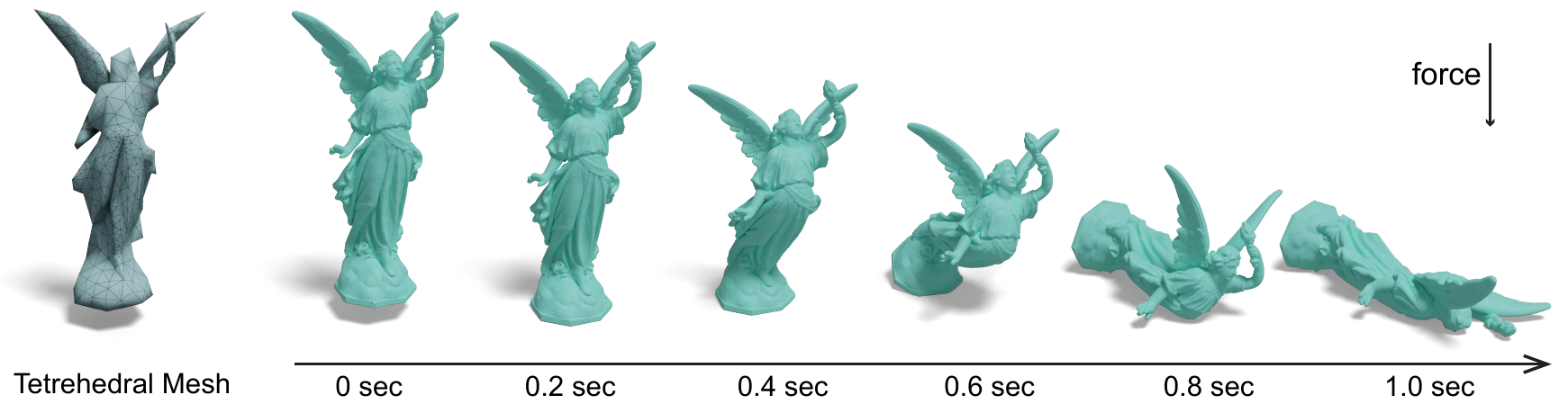}
  \end{center}
   \figspace
  \caption{\textbf{The statue collides with the ground and deforms.}  Our implicit neural representation (green, right)
  is capable of capturing more fine geometry details compared to the traditional
  tetrahedral mesh representation (blue, left) under the same memory footprint. }
  \label{3d_lucy_collide_plane}
  \vspace{-10pt}
\end{figure*}

\subsection{Elastodynamic Equation}
\label{app:bc_elasticity}

\paragraph{Initial and Boundary Conditions}

For our 2D elasticity examples in \Cref{2d_patch_test_vs_fem_mpm} and \Cref{2d_square_collide_sphere}, the 2D spatial domain is $\spatialDomain=[-1, 1]\times[-1, 1]$. 
For our 3D elasticity examples in \Cref{3d_elasticity_convergence}, the 3D spatial domain is $\spatialDomain=[-1, 1]\times[-1, 1]\times[-1, 1]$. For our 2D and 3D examples involving nonregular geometry (\Cref{3d_bunny_collide_plane}, \Cref{fig:opti-time-inte-nn} and \Cref{3d_lucy_collide_plane}), the spatial domain is the interior of the shape, including the boundary.
The initial condition for all the elasticity examples is 
\begin{align}
\begin{split}
	\hat{\deformMap^0}(\pos) =(0, 0) & \text{ (2D)}, \\
  \hat{\deformMap^0}(\pos) =(0, 0, 0) & \text{ (3D)}
\end{split}
\end{align}

The boundary constraint for elasticity examples involves positional constraints or collision constraints. Positional constraints, or \textit{Dirichlet boundary conditions}, can be realized by defining the position of the constraint set $\partial\spatialDomain$ as the desired goal positions $\overline\deformMap_{\partial\spatialDomain}$:
\begin{align}
  \TimeIntegration_{\text{pos}}&=\squaredNorm{\defomMapDiscretePlus_{\partial\spatialDomain} - \overline\deformMap_{\partial\spatialDomain}}.
\end{align}

Collision constraints can be handled by adding unilateral constraints dynamically and viewing the collision penalty force as an external force.
Specifically, for a colliding point $\qb_c$, we first find the closest surface point $\bb_c$ with normal $\nb_c$, and define our spring-like collision penalty force as:
\begin{align}
  \externalforce_{\text{col}}=k_{\text{col}} ((\bb_c-\qb_c)^\top \nb_c) \nb_c.
\end{align}
where $k_{\text{col}}$ is the ratio for the collision penalty force.

The corresponding collision energy can be defined as the work exerted by the collision force:
\begin{align}
  \TimeIntegration_{\text{col}} =\solidDen \externalforce_{\text{col}}^T\defomMapDiscretePlus.
\end{align}

\paragraph{Experiment Setup}

For all the 2D comparisons under the same memory usage, we use $\alpha = 3$ hidden layers of width $\beta = 68$ with SIREN activation function \citep{sitzmann2020implicit} for our MLP, which takes the same memory (57 KB) as the FEM mesh (0.8K vertices, 1.5K faces) and MPM point cloud (1.7K points) in use. 
We initialize the 2D deformation field of the network to be zero using $|\sampleSet| = 1000^2$ uniform and random samples. Then we train the network using $|\sampleSet| = 100^2$ uniform and random samples at each training iteration.
We use Bartels \citep{levin2020bartels} and Taichi \citep{hu2019taichi} to perform the FEM and MPM simulation, respectively. We run our FEM and MPM comparison on CPU using a MacBook Pro with the Apple M2 processor and 24GB of RAM.

For the 3D comparison under the same memory usage, for the bunny example (\Cref{3d_bunny_collide_plane}), we use $\alpha = 3$ hidden layers of width $\beta = 66$ with SIREN activation function for our MLP, which takes the same memory (53 KB) as the FEM mesh (0.5K vertices, 1.5K tetrahedra) in use. For the statue example (\Cref{3d_lucy_collide_plane}), we use $\alpha = 3$ hidden layers of width $\beta = 128$ with SIREN activation function for our MLP, which takes the same memory (197 KB) as the FEM mesh (2.0K vertices, 7.0K tetrahedra) in use. 
We initialize the 3D deformation field of the network to be zero using $|\sampleSet| = 100^3$ uniform and random samples. 
Then we train the network using $|\sampleSet| = 20^3$ uniform and random samples at each training iteration. Here for simplicity, we use the mesh vertices as the uniform samples.
We further report all the parameters and experiment setup in \Cref{tbl:elasticity_setup}. 
In addition, we set the hyper-parameters $\iterPatience=800$ and $\lrMin=1e{-8}$ for all elasticity examples and assume a constant density in the reference space.

The geometry (i.e., the undeformed shape) can be any representation (e.g., analytical, mesh, or level set), as long as it allows sampling within the volume. For our examples in \Cref{2d_patch_test_vs_fem_mpm} and \Cref{2d_square_collide_sphere}, the geometry (a square) is represented analytically. For examples in \Cref{3d_bunny_collide_plane} and \Cref{3d_lucy_collide_plane}, the geometry is represented using the original high-resolution mesh.

For sampling of the shapes involving nonregular geometry, for simplicity, we choose to use a triangle or tetrahedral mesh and perform sampling within the volume \emph{during the training}. An ideal alternative would be adopting the implicit representation of the surface and performing rejection sampling based on it.

For rendering, we sample a sufficient number of points from the undeformed shape and evaluate the trained model at time $t$ on the sample positions to predict their deformation. Thus the deformed shape $x^t$ at each time step $t$ can be obtained by applying the deformation field $\deformMap^t$ on the sample points of undeformed shape $x^0$, i.e., $x^t = x^0 + \deformMap^t(x^0)$. For visualization, we only sample the surface of the shape in 3D cases. Then we render the shape as a dense point cloud. 

\vspace{-3pt}
\section{Additional Results}
\label{app:add_results}

\subsection{Elastodynamic Equation}

In \Cref{3d_twisting_test}, we demonstrate that our method exhibits volume-preserving property on a 3D twisting example.
In \Cref{3d_lucy_collide_plane}, we provide another example involving complex contact-induced deformations.

The L2 distance errors in \Cref{2d_patch_test_error} and \Cref{tension_quant} are computed using the vertices of the reference FEM mesh.
Specifically, the L2 distance is defined between the deformed positions of those vertices in the reference solution and in our/compared FEM solution.
In our solution, the deformed positions of those vertices are queried via direct network inference.
In the compared low-resolution FEM solution, the deformed positions of those vertices are calculated using the barycentric interpolation.

Finally, we demonstrate our method's qualitative and quantitative convergence when increasing the number of spatial samples for training. In \Cref{3d_elasticity_convergence}, we compare the quasistatic stretching results when using a different number of spatial samples (recall \Cref{para:sampling} Spatial Sampling), and report the error with respect to the reference result (using $|\sampleSet|=50^3$).


\end{document}